\documentclass{article}

\usepackage{microtype}
\usepackage{graphicx}
\usepackage{subfigure}
\usepackage{booktabs} 

\usepackage{hyperref}

\usepackage[accepted]{icml2021}

\usepackage{amsthm}
\newtheorem{definition}{Definition}
\newtheorem{theorem}{Theorem}
\newtheorem{corollary}{Corollary}[theorem]
\newtheorem{lemma}[theorem]{Lemma}

\newtheorem{postulate}[theorem]{Postulate}
\theoremstyle{remark}
\newtheorem*{remark}{Remark}

\usepackage{graphicx}
\usepackage{caption}
\usepackage{subfigure}

\usepackage{epigraph}

\usepackage{xparse}
\ExplSyntaxOn

\makeatletter
\NewDocumentCommand{\multicitep}{m}
 {
  \NAT@open
  \mjb_multicitep:n { #1 }
  \NAT@close
 }
\makeatother

\seq_new:N \l_mjb_multicite_in_seq
\seq_new:N \l_mjb_multicite_out_seq
\seq_new:N \l_mjb_cite_seq

\cs_new_protected:Npn \mjb_multicitep:n #1
 {
  \seq_set_split:Nnn \l_mjb_multicite_in_seq { ; } { #1 }
  \seq_clear:N \l_mjb_multicite_out_seq
  \seq_map_inline:Nn \l_mjb_multicite_in_seq
   {
    \mjb_cite_process:n { ##1 }
   }
  \seq_use:Nn \l_mjb_multicite_out_seq { ;~ }
 }

\cs_new_protected:Npn \mjb_cite_process:n #1
 {
  \seq_set_split:Nnn \l_mjb_cite_seq { , } { #1 }
  \int_compare:nTF { \seq_count:N \l_mjb_cite_seq == 1 }
   {
    \seq_put_right:Nn \l_mjb_multicite_out_seq
     { \citeauthor{#1},~\citeyear{#1} }
   }
   {
    \seq_put_right:Nx \l_mjb_multicite_out_seq
     {
      \exp_not:N \citeauthor{\seq_item:Nn \l_mjb_cite_seq { 1 }},~
      \exp_not:N \citeyear{\seq_item:Nn \l_mjb_cite_seq { 1 }},~
      \seq_item:Nn \l_mjb_cite_seq { 2 }
     }
   }
 }
\ExplSyntaxOff

\newcommand{\indep}{\perp \!\!\! \perp}

\newcommand{\modelacronym}{ACML}
\newcommand{\societyacronym}{SDM}
\newcommand{\clonedvickreysociety}{CVS}

\usepackage{amssymb}
\usepackage{lipsum}
\usepackage{amsmath}

\newcommand{\given}{\;|\;}

\newcommand{\mgiven}{\;\middle|\;}

\newcommand{\camech}{\Pi}
\newcommand{\alldata}{\textbf{x}}
\newcommand{\trace}{\textbf{x}}
\newcommand{\allfunc}{\textbf{\texttt{f}}}
\newcommand{\compgraph}{\textbf{\texttt{G}}}
\newcommand{\mexec}{\textbf{\texttt{E}}}
\newcommand{\metamdp}{\textbf{\texttt{C}}}
\newcommand{\acl}{\mathbb{L}}

\newcommand{\learnindex}{i}
\newcommand{\numfunc}{N}
\newcommand{\tracelen}{T}
\newcommand{\traceindex}{t}
\newcommand{\othertraceindex}{{\neq \traceindex}}
\newcommand{\funcindex}{k}
\newcommand{\otherfuncindex}{j}
\newcommand{\thisfunc}{\texttt{f}^\funcindex}
\newcommand{\otherfunc}{\texttt{f}^\otherfuncindex}
\newcommand{\thisfeedback}{\delta^\funcindex}

\usepackage[dvipsnames]{xcolor}

\usepackage{bm}

\usepackage{tikz}
\usepackage[framemethod=TikZ]{mdframed}
\mdfdefinestyle{MyFrame2}{%
    linecolor=white,
    outerlinewidth=1pt,
    roundcorner=2pt,
    innertopmargin=\baselineskip,
    innerbottommargin=\baselineskip,
    innerrightmargin=10pt,
    innerleftmargin=10pt,
    backgroundcolor=black!3!white}
    
\surroundwithmdframed[
  leftmargin=1pt,
  skipabove=\medskipamount,
  skipbelow=\medskipamount
]{example}

\usepackage{etoc}
\usepackage{appendix}

\newcommand\blfootnote[1]{%
  \begingroup
  \renewcommand\thefootnote{}\footnote{#1}%
  \addtocounter{footnote}{-1}%
  \endgroup
}

\icmltitlerunning{Modularity in Reinforcement Learning via Algorithmic Independence in Credit Assignment}

\begin{document}

\twocolumn[
\icmltitle{Modularity in Reinforcement Learning\\via Algorithmic Independence in Credit Assignment}



\icmlsetsymbol{equal}{*}

\begin{icmlauthorlist}
\icmlauthor{Michael Chang*}{1}
\icmlauthor{Sidhant Kaushik*}{1}
\icmlauthor{Sergey Levine}{1}
\icmlauthor{Thomas L. Griffiths}{2}
\end{icmlauthorlist}

\icmlaffiliation{1}{Department of Computer Science, University of California, Berkeley, USA}
\icmlaffiliation{2}{Department of Computer Science, Princeton University, USA}

\icmlcorrespondingauthor{Michael Chang}{mbchang@berkeley.edu}

\icmlkeywords{Reinforcement Learning, Algorithmic Information Theory, Causality, Credit Assignment, Modularity}

\vskip 0.3in
]

\blfootnote{ICML (long) oral presentation: \url{https://sites.google.com/view/modularcreditassignment}.}



\printAffiliationsAndNotice{\icmlEqualContribution} 

\begin{abstract}
Many transfer problems require re-using previously optimal decisions for solving new tasks, which suggests the need for learning algorithms that can modify the mechanisms for choosing certain actions independently of those for choosing others. However, there is currently no formalism nor theory for how to achieve this kind of modular credit assignment. To answer this question, we define modular credit assignment as a constraint on minimizing the algorithmic mutual information among feedback signals for different decisions. We introduce what we call the modularity criterion for testing whether a learning algorithm satisfies this constraint by performing causal analysis on the algorithm itself. We generalize the recently proposed societal decision-making framework as a more granular formalism than the Markov decision process to prove that for decision sequences that do not contain cycles, certain single-step temporal difference action-value methods meet this criterion while all policy-gradient methods do not. Empirical evidence suggests that such action-value methods are more sample efficient than policy-gradient methods on transfer problems that require only sparse changes to a sequence of previously optimal decisions.
\end{abstract}
\epigraph{It is causality that gives us this modularity, and when we lose causality, we lose modularity.}{Judea Pearl~\citep{ford2018architects}}

\begin{figure}[!b]
    \centering
    \includegraphics[width=0.45\textwidth]{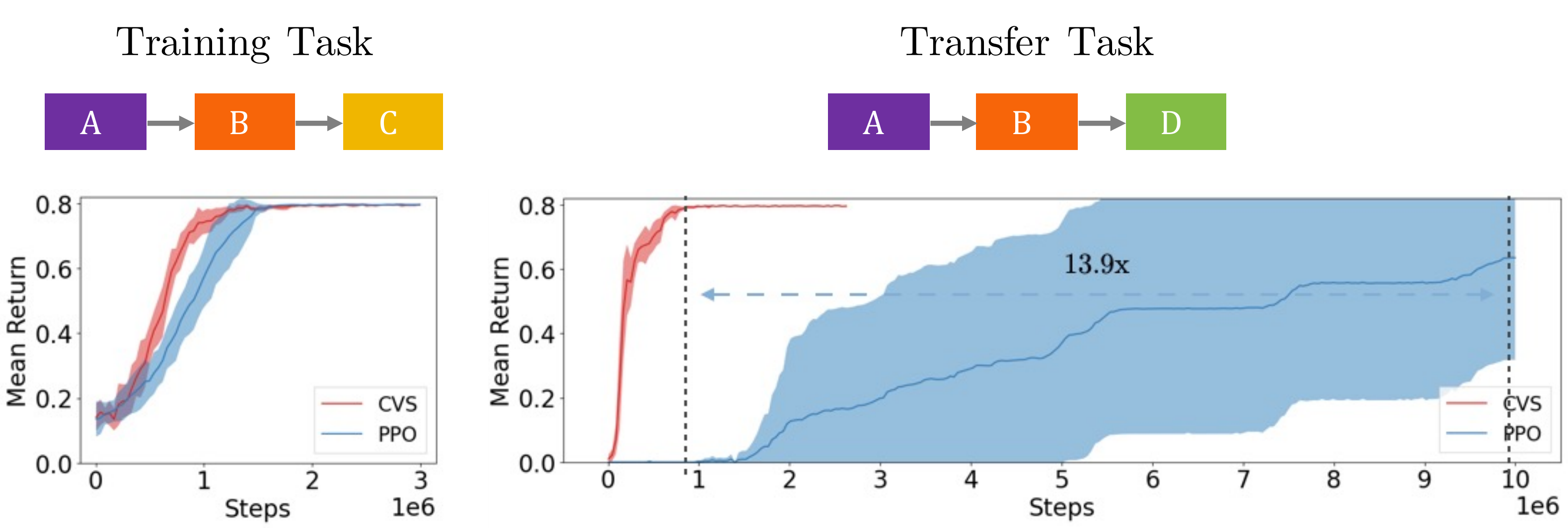}
    \caption{\small{
    \textbf{Minimal motivating example.}
    The optimal action sequence for the training task is $A \rightarrow B \rightarrow C$, and the optimal sequence for the transfer task differs only in the last time-step.
    Continuing to train an optimal policy from the training task on the transfer task with the cloned Vickrey society (\clonedvickreysociety) from~\citet{chang2020decentralized} transfers $13.9$x more efficiently than with PPO~\citep{schulman2017proximal}, even though learning efficiency for both on-policy algorithms during training is comparable.
    This paper suggests that this is due to \textbf{dynamic modularity}: the algorithmic independence among~\clonedvickreysociety's learnable mechanisms and among their gradients.
    }}
    \label{fig:motivating_example}
\end{figure}

\section{Introduction} \label{sec:introduction}
Gusteau's~\citep{Ratatouille2007} taqueria has a great team for making burritos: Colette heats the tortillas, Remy adds the meat, and Alfredo wraps the burrito in aluminum foil.
But today customers fell sick from meat contamination and gave angry reviews.
Clearly, Remy should replace meat with tofu or something else.
But should credit assignment from the reviews affect the others?
Intuitively, no: the feedback signals to the decision of adding meat and to the decisions of heating tortillas and wrapping aluminum foil should be independent.
Customer dissatisfaction in burritos are not reflective of the taqueria's quesadillas, for which Colette's tortilla skills and Alfredo's wrapping skills are useful.

The example above expresses the intuition that \textbf{modularity}, or the capacity for the mechanisms in a system to be independently modified, enables flexible adaptation.
However, using principles of modularity to build flexible learning systems has been difficult because the traditional formalism for precisely describing what modularity means was developed in the context of analyzing \textbf{static systems} -- systems whose \textbf{mechanisms}, or functional components, are assumed fixed.
But learning agents are \textbf{dynamic systems} composed of mechanisms (i.e. learnable functions) that evolve over the course of learning.
Thus, to even express the hypothesis that modularity enables flexibility in learning agents, let alone test it, we first need \textcolor{RedOrange}{(1)} a formalism that defines what modularity means for dynamic systems, \textcolor{OliveGreen}{(2)} a theory that 
identifies the conditions under which independent modification of learnable mechanisms is even possible, and \textcolor{NavyBlue}{(3)} a practical criterion for determining when these conditions are met in learning algorithms.
This paper proposes candidate solutions to these problems and applies them to shed new insight on the modularity of discrete-action reinforcement learning (RL) algorithms.
The takeaway message of this paper is that independent modification of mechanisms requires both the mechanisms \emph{and} the feedback signals that update them to be independent: a modular learning algorithm must have a credit assignment mechanism whose algorithmic causal structure makes such independent modification possible.

\citet{janzing2010causal} proposed to precisely characterize modularity in static systems as the algorithmic independence of mechanisms in the computational graph used to describe the system.
In learning systems, the computational graph in question depicts the forward pass of a learner (e.g. a neural network), but this graph itself evolves over the course of learning because the learnable functions -- the mechanisms -- get modified.
For such dynamic systems, we extend the static notion of modularity to define \textbf{dynamic modularity} as the algorithmic independence of mechanisms in the current iteration, conditioned on the graph from the previous iteration of evolution.
This addresses problem \textcolor{RedOrange}{(1)}.

Modularity matters when the system needs to be modified for a new context or purpose.
In learning systems it is the credit assignment mechanism that performs this modification.
Thus dynamic modularity is tied to independence in feedback: for a gradient-based learner, we show that enforcing dynamic modularity requires enforcing gradients to be algorithmically independent as well, which we call the \textbf{modularity constraint.}
This addresses problem \textcolor{OliveGreen}{(2)}.
 
Algorithmic independence is generally incomputable, which makes the modularity constraint intractable to evaluate.
To make this constraint practical for analyzing learning algorithms, we formally represent the entire learning process as one big causal graph, which we call the \textbf{algorithmic causal model of learning} (\modelacronym).
Then the modularity constraint translates into an easy-to-inspect criterion, the \textbf{modularity criterion}, on $d$-separation in~\modelacronym~that enables us to evaluate, without any training, whether a learning algorithm is exhibits dynamic modularity.
This addresses problem \textcolor{NavyBlue}{(3)}.

Having established a theoretically-grounded formalism for reasoning about modularity in learning systems, we theoretically and empirically analyze discrete-action RL algorithms.
The mechanisms of interest are the functions that compute the ``bid'' (e.g. action probability or $Q$-value) for each value of the action variable.
The Markov decision process (MDP) is too coarse-grained to represent these functions separately, so we use the societal decision-making framework (\societyacronym) from~\citet{chang2020decentralized}, whose computational graph does treat them separately.
We prove that certain single-step temporal difference methods satisfy the modularity criterion while all policy gradient methods do not.
Empirically, we find that for transfer problems that require only sparse modifications to a sequence of previously optimal decisions, implementations of algorithms that exhibit dynamic modularity transfer more efficiently than their counterparts.
All proofs are in the Appendix.

\paragraph{Assumptions and approximations}\label{sec:assumptions_and_approximations}
The theory developed in~\S\ref{sec:dynamic_modularity_in_learning_systems},~\S\ref{sec:an_algorithmic_causal_model_of_learning}, and~\S\ref{sec:modularity_in_reinforcement_learning} assume computational graphs over arbitrary strings.
Thus we will understand statements about Kolmogorov complexity and algorithmic information in these sections as inherently asymptotic, pertaining to strings of increasing length, where equations that hold up to constant terms are well defined.
Similarly, our discussion in~\S\ref{sec:modularity_in_reinforcement_learning} will pertain not to concrete instantiations of RL algorithms but only to the causal structure of these algorithms in the abstract.
As with all explanations in science, there is inevitably a gap between theory and practice.
In particular, when considering empirical performance of concrete instantiations of RL algorithms in~\S\ref{sec:experiments} on a specific Turing machine, asympototic statements are not meaningful, but we can still use empirical observation to refute or improve our theory.
\section{Related Work} \label{sec:related_work}
The hypothesis that modularity could improve flexibility of learning systems has motivated much empirical work in designing factorized architectures~\citep{devin2017learning,andreas2016neural,chang2018automatically,goyal2019recurrent,kirsch2018modular,alet2018modular,pathak2019learning} and reinforcement learners~\citep{simpkins2019composable,sprague2003multiple,samejima2003inter}, but the extent to which the heuristics used in these methods enforce the learnable components to be independently modifiable has yet to be tested.
Conversely, other works begin by defining a multi-agent system of independently modifiable components and seek methods to induce their cooperation 
with respect to a global objective~\citep{balduzzi2014cortical,baum1996toward,srivastava2013compete,chang2020decentralized,gemp2020d3c,balduzzi2020smooth}, but the precise property of a learning system that characterizes its modularity has not been discussed in these works, as far as we are aware.
Recent complementary work has proposed alternative measures of modularity, restricted to deep networks, based on connectivity strength~\citep{filan2021clusterability} and functional decomposition~\citep{csordas2020neural}.
In contrast, our work identifies a general property that defines the modularity of a learning system as the algorithmic independence of learnable mechanisms and of their gradients, and presents a practical method for testing for this property without any training.
We build upon the theoretical foundations from~\citet{janzing2010causal} that have clarified similar notions of ``autonomy'' and ``invariance'' that underlie axioms of econometrics~\citep{haavelmo1944probability,aldrich1989autonomy}, causality~\citep{pearl2009causality,peters2017elements}, and computer programming~\citep{abelson1996structure}.
\citet{yu2020gradient} explored enforcing the linear independence of gradients to improve multi-task learning, and formulating the precise connection between algorithmic and linear independence would be valuable future work.
\section{Background} \label{sec:preliminaries}
Our analysis of the modularity of RL algorithms employs two key ideas: (\S\ref{sec:algorithmic_causality}) computational graphs can be interpreted as causal graphs and (\S\ref{sec:societal_decision_making}) a learnable discrete-action policy can be interpreted as a society of learnable action-specific functions and a fixed selection mechanism.

\subsection{Algorithmic Causality} \label{sec:algorithmic_causality}
We begin by reviewing terms from algorithmic information theory~\citep{kolmogorov1965three,li2008introduction,solomonoff1964formal}\footnote{See the appendix for background.}.
We assume that programs are expressed in a language $L$ and run on a universal Turing machine.
Given binary strings $x$, $y$, and $z$, we denote \textbf{conditional algorithmic independence} as $x \indep y \given z$, equivalently $I\left(x : y \mgiven z\right) \overset{+}{=} 0$, which reads ``given $z$, knowledge of $y$ does not allow for a stronger compression of $x$.''
Let $x^*$ denote the shortest program that produces $x$.
$I$ denotes \textbf{conditional algorithmic mutual information}.
$\overset{+}{=}$ denotes equality up to a constant that depends on $L$ but not the strings on either side of the equality (see~\S\ref{sec:assumptions_and_approximations}).
\textbf{Conditional Kolmogorov complexity} of $y$ given $x$ is given by $K(y \given x)$, the length of the shortest program that generates $y$ from $x$ as input.

\citet[Post.~6]{janzing2010causal} generalized structural causal models~\citep{pearl1995causal} to general programs, allowing us to treat computational graphs as causal graphs.
\begin{definition}[\textbf{computational graph}] \label{def:computational_graph}
Define a \textbf{computational graph} $\compgraph = (\alldata, \allfunc)$ as a directed acyclic factor graph (DAG) of variable nodes $\alldata = x_1, ..., x_N$ and function nodes $\allfunc = \texttt{f}^1, ..., \texttt{f}^N$.
Let each $x_j$ be computed by a program \texttt{f}$^j$ with length $O(1)$ from its parents $\{pa_j\}$ and an auxiliary input $n_j$.
Assume the $n_j$ are jointly independent: $n_j \indep \{n_{\neq j}\}$.
Formally, $x_j := \texttt{f}^j(\{pa_j\}, n_j)$, meaning that the Turing machine computes $x_j$ from the input $\{pa_j\}, n_j$ using the additional program $\texttt{f}^j$ and halts.
\end{definition}
By absorbing the $n_j$ into the functions $\texttt{f}^j$ we can equivalently assume that $\texttt{f}^j$ are jointly independent, but not necessarily $O(1)$~\citet[Post.~6]{janzing2010causal}.
If the $n_j$ are interpreted as noise, this DAG represents a probabilistic program~\citep{van2018introduction,probmods2,mansinghka2009natively} that implements a standard causal model.
Henceforth we treat all graphs as computational graphs.
We define a \textbf{mechanism} as the string representation (i.e. source code) of the program that implements a function $\texttt{f}$ and \textbf{data} as the string representations of the input/output variables $x$ of $\texttt{f}$.
The \textbf{algorithmic causal Markov condition}~\citep[Thm.~4]{janzing2010causal}, which states that $d$-separation implies conditional independence, generalizes the standard Markov condition to general programs:
\begin{theorem}[\textbf{algorithmic causal Markov condition}] \label{thm:algorithmic_causal_markov_condition}
Let $\{pa_j\}$ and $\{nd_j\}$ respectively represent concatenation of the parents and non-descendants (except itself) of $x_j$ in a computational graph.
Then $\forall x_j$, $x_j \indep \{nd_j\} \given \{pa_j\}^*.$
\end{theorem}
In standard causality it is typical to assume the converse of the Markov condition, known as \textbf{faithfulness}~\citep{spirtes2000causation}.
We do the same for algorithmic causality:
\begin{postulate}[\textbf{algorithmic faithfulness}] \label{post:faithfulness}
Given sets $S$, $T$, $R$ of nodes in a computational graph, $I(S : T | R^*) \overset{+}{=} 0$ implies $R$ d-separates $S$ and $T$.
\end{postulate}

\subsection{Societal Decision-Making} \label{sec:societal_decision_making}
A discrete-action MDP is the standard graph for a sequential decision problem over states $S$ with $N$ discrete actions $A$, defined with state space $\mathcal{S}$, action space $\{1, ..., N\}$, transition function $\texttt{T}: \mathcal{S} \times \{1, ..., N\} \rightarrow \mathcal{S}$, reward function $\texttt{R}: \mathcal{S} \times \{1, ..., N\} \rightarrow \mathbb{R}$, and discount factor $\gamma$.
The MDP objective is to maximize the return $\sum_{t=0}^T \gamma^t \texttt{R}(s_t, a_t)$ with respect to a policy $\pi: \mathcal{S} \rightarrow \{1, ..., N\}$.
We define a \textbf{decision} as a value $a$ of $A$.
The MDP abstracts over the mechanisms that control each decision with a single edge in the graph, represented by $\pi$, but to analyze the independence of different decisions we are interested in representing these mechanisms as separate edges.

The societal decision-making (\societyacronym) framework~\citep{chang2020decentralized} offers an alternative graph that does exactly this: it decomposes a discrete-action policy as a society of $\numfunc$ agents $\omega^\funcindex$ that each controls a different decision.
Each agent is a tuple $(\psi^\funcindex, \phi^\funcindex)$ of a bidder $\psi^\funcindex: \mathcal{S} \rightarrow \mathcal{B}$ and a fixed transformation $\phi^\funcindex: \mathcal{S} \rightarrow \mathcal{S}$.
In~\S\ref{sec:modularity_in_reinforcement_learning}, we will consider the algorithmic independence of the $\psi^n$.
Recovering a policy $\pi$ composes two operations: one computes bids $b^\funcindex_s := \psi^\funcindex(s),\;\forall \funcindex$, and one applies a selection mechanism $\texttt{S}: \mathcal{B}^\numfunc \rightarrow \{1, ..., N\}$ on the bids to select decision $a$.
\societyacronym~thus curries the transition and reward functions as $\texttt{T}: \{1, ..., N\} \rightarrow [\mathcal{S} \rightarrow \mathcal{S}]$ and $\texttt{R}: \{1, ..., N\} \rightarrow [\mathcal{S} \rightarrow \mathbb{R}]$.

\citet{chang2020decentralized} introduced the cloned Vickrey society (\clonedvickreysociety) algorithm as an on-policy single-step temporal-difference action-value method.
\clonedvickreysociety~interprets the Bellman optimality equation as an economic transaction between agents seeking to optimize their utilites in a Vickrey auction~\citep{vickrey1961counterspeculation} at each time-step.
The Vickrey auction is the selection mechanism that selects the highest bidding agent $i$, which receives a utility
\begin{equation} \label{eqn:sdm_utility}
    \underset{\text{utility}}{\underbrace{\vphantom{\max_{j \neq i}}U^i_{s_t}(\omega^{1:N})}} = \underset{\text{revenue, or valuation} \; v_{s_t}}{\underbrace{\vphantom{\max_{j \neq i}}\texttt{R}^\phi(\omega^i, s_t) + \gamma \cdot \max_k b_{s_{t+1}}^k}} - \underset{\text{price}}{\underbrace{\max_{j \neq i} b_{s_t}^j}},
\end{equation}
and the rest receive a utility of $0$.
In~\clonedvickreysociety~each agent bids twice: the highest and second highest bids are produced by the same function parameters.
The auction incentivizes each agent to truthfully bid the $Q$-value of its associated transformation mechanism, independent of the identities and bidding strategies of other agents.

\begin{figure*}
    \centering
    \includegraphics[width=\textwidth]{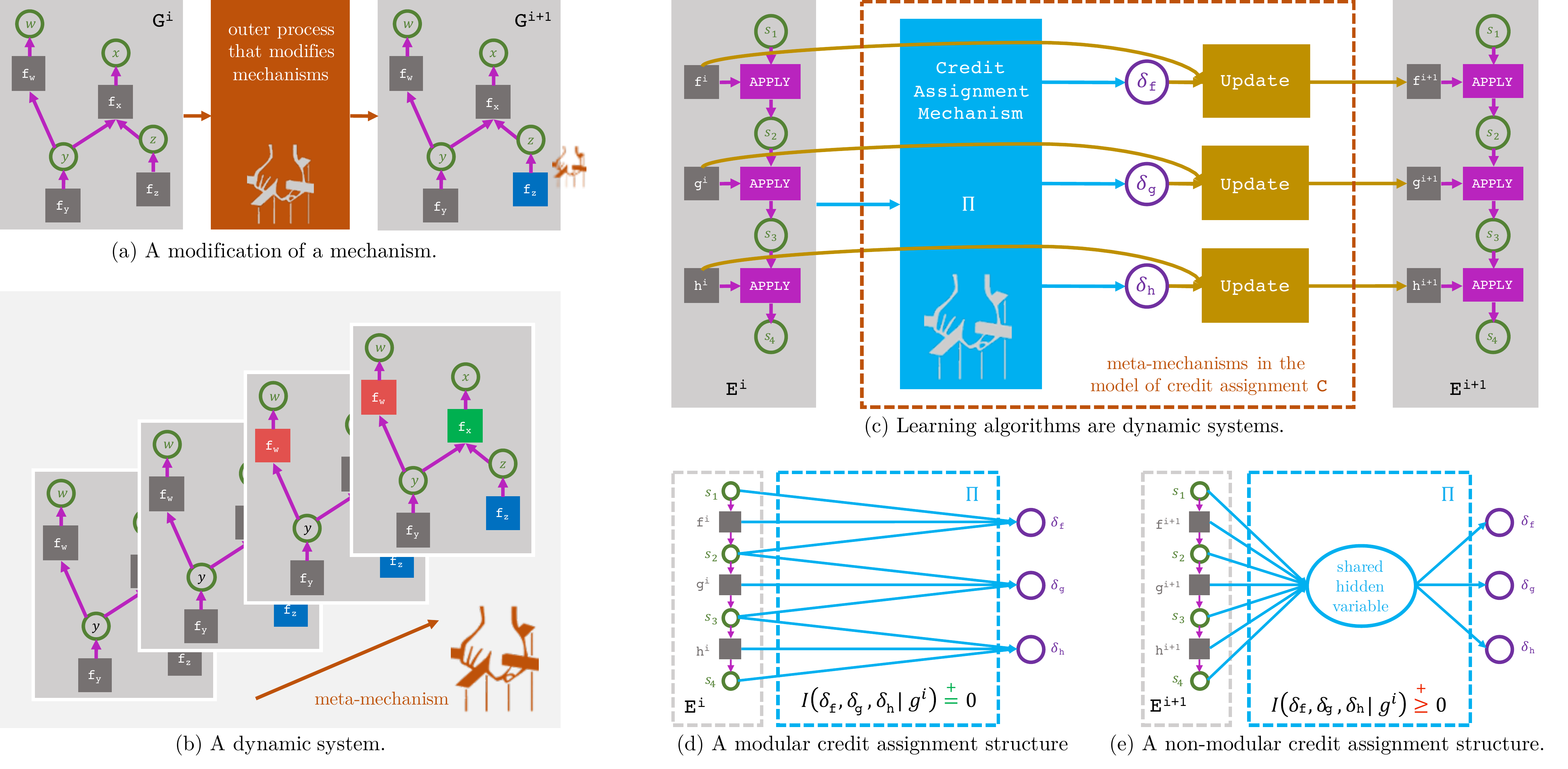}
    \caption{
    \small{\textbf{Key Ideas.}
    A system can be represented as a algorithmic causal graph $\texttt{G}$.
    (a) 
    A modification to a mechanism in graph $\texttt{G}^i$ generates a new graph $\texttt{G}^{i+1}$.
    (b) A dynamic system encompasses an outer process that generates a sequence of graphs via a series of modifications to the mechanisms.
    (c) Learning algorithms are examples of dynamic systems, where the outer process is the model of credit assignment $\texttt{C}$, which modifies the mechanisms of the model of execution $\texttt{E}$, which represents the forward pass of the learner.
    By flattening the learning algorithm as one algorithmic causal graph, we can determine whether the causal structure of the credit assignment mechanism makes independent modification of learnable mechanisms possible by inspecting whether the gradients are $d$-separated by the previous graph $\texttt{C}^i$.
    A credit assignment mechanism is (d) modular if they are $d$-separated and (e) not modular if not.
    }}
    \label{fig:main}
\end{figure*}

\section{Dynamic Modularity in Learning Systems} \label{sec:dynamic_modularity_in_learning_systems}
This section extends the definition of modularity in static systems to dynamic systems.
We discuss learning algorithms as examples of dynamic systems and the constraints that must be imposed on the credit assignment mechanism for the learning algorithm to be exhibit dynamic modularity.

\subsection{From Static Modularity to Dynamic Modularity} \label{sec:from_static_modularity_to_dynamic_modularity}
A system can be described by a computational graph (i.e. an algorithmic causal graph), whose mechanisms represent the system components and whose data represent the information communicated between components.
In standard causal analysis~\citep{pearl2009causality}, mechanisms are generally treated as fixed.
In this context, we call the graph a \textbf{static graph} that describes a \textbf{static system.}
Modularity in the static context has been defined as the autonomy~\citep[\S1.3.1]{pearl2009causality}, or more precisely, the algorithmic independence~\citep[Post.~7]{janzing2010causal} of mechanisms:
\begin{definition}[\textbf{static modularity}] \label{eqn:static_modularity}
\begin{align}
    \forall \funcindex \neq \otherfuncindex, \quad I\left(\thisfunc : \otherfunc\right) \overset{+}{=} 0.
\end{align}

\end{definition}
If mechanisms are algorithmically independent, then one can be modified without an accompanying modification in the others to compensate.
This enables the analysis of counterfactual queries, for example, where a human performs a hypothetical modification to a mechanism of a static graph (Fig.~\ref{fig:main}a).
To analyze a different query, the mechanism is reset to its original state before the hypothetical modification, and the human then performs a different hypothetical modification on the original static graph.
The human is a \textbf{meta-mechanism}, a mechanism that modifies mechanisms.

Whereas static graphs are useful for analyzing systems under human control, many systems in the real world (e.g. star systems, whose mechanisms are stars that communicate via forces) are \textbf{dynamic systems} (Fig.~\ref{fig:main}b), whose mechanisms evolve through time (e.g. the stars move).
We describe these systems with a \textbf{dynamic graph}.
In the dynamic context, the meta-mechanism is not the human, but symmetric laws that govern the time evolution of the mechanisms (e.g. physical laws governing stars' motion are invariant to change reference frame).
Since any snapshot in time of a dynamic system depicts a static system, we naturally extend the traditional definition of modularity to the dynamic context:
\begin{definition}[\textbf{dynamic modularity}] \label{eqn:dynamic_modularity}
\begin{align}
    \forall \funcindex \neq \otherfuncindex, \quad I\left(\texttt{f}^{\funcindex,\learnindex+1} : \texttt{f}^{\otherfuncindex,\learnindex+1} \mgiven \trace^\learnindex, \allfunc^\learnindex \right) \overset{+}{=} 0.
\end{align}
\end{definition}
Dynamically modularity simply re-interprets static modularity as a snapshot $i$ along the temporal dimension.

\subsection{Learning Algorithms are Dynamic Systems} \label{sec:learning_algorithms_are_dynamic_systems}
We now show that general learning algorithms are examples of dynamic systems and can be analyzed as such.
To do so, we need to specify the data and mechanisms of the static computational graph that represents a particular snapshot, as well as the equivariant meta-mechanism that evolves the mechanisms from one iteration to the next.

Let the \textbf{model of execution} be the computational graph $\mexec$ that represents the forward pass of the learner, generating $\trace$ as an execution trace $(x_1, ..., x_\traceindex, ..., x_\tracelen)$ of the input and output data of the learnable mechanisms $\allfunc$.
For example, with MDPs, the forward pass is a rollout, the trace records its states, actions, and rewards, and the mechanisms, which map parent variables $\{pa\}_\traceindex = s_\traceindex$ to child variables $x_\traceindex = (a_\traceindex, s_{\traceindex+1}, r_\traceindex)$, are instances of the policy at different steps $\traceindex$.

Let the \textbf{model of credit assignment} be the computational graph $\metamdp$ that evolves the mechanisms.
Each step represents the backward pass of the learner.
Here the mechanisms are treated as data for two equivariant meta-mechanisms, the credit assignment mechanism $\camech(\trace, \allfunc) \rightarrow \bm{\delta}$ and the update rule $\texttt{UPDATE}(\allfunc, \bm{\delta}) \rightarrow \allfunc'$.
$\metamdp$ can be viewed as a reward-less MDP with states $\allfunc$ and actions $\bm{\delta}$, with $\texttt{UPDATE}$ as the transition function.
Then $\camech$ is a context-conditioned policy that generates modifications $\bm{\delta} = (\delta_1, ..., \delta_\tracelen)$ to the functions $\allfunc$ of the learner, given $\trace$ as context.
For a gradient-based learner, $\thisfeedback_\traceindex$ is the gradient of the learning objective with respect to the function $\thisfunc$ that participated at step $\traceindex$ of the execution trace (e.g. as we discuss in~\S\ref{sec:modularity_in_reinforcement_learning}, $\thisfeedback_\traceindex$ would be the Bellman error of the decision mechanism for action $k$ taken at step $i$).
\texttt{UPDATE} performs the parallel operation $\texttt{UPDATE}(\thisfunc, \sum_{\traceindex} \thisfeedback_\traceindex) \rightarrow \texttt{f}^{\funcindex\prime}$ over all mechanisms $\thisfunc$.
The choice of optimizer for gradient descent (e.g. Adam~\citep{kingma2014adam}) determines the functional form of \texttt{UPDATE}.
Henceforth we assume gradient-based learning, but our results hold more generally given the assumptions that $\texttt{UPDATE}$ (1) is algorithmically independent of $\camech$ and (2) completely factorizes across $\funcindex$.

\subsection{Modularity Constraint on Credit Assignment} \label{sec:modularity_constraint}
The design of a learning algorithm primarily concerns the credit assignment mechanism $\camech$, whereas the choice of \texttt{UPDATE} is often assumed.
We now present the constraint $\camech$ must satisfy for dynamic modularity to hold at every iteration of learning.
Given trace $\trace$ and previous mechanisms $\allfunc$, we define the \textbf{modularity constraint} as that which imposes that the gradients $\delta_1, ..., \delta_\tracelen$ be jointly independent:
\begin{definition}[\textbf{modularity constraint}] \label{def:modularity_constraint}
\begin{align}
    I\left(\delta_1, ..., \delta_\tracelen \mgiven \trace, \allfunc \right) &\overset{+}{=} 0.
\end{align}
\end{definition}
A \textbf{modular credit assignment mechanism} is one that satisfies the modularity constraint.
If $\mexec$ exhibited statically modularity (i.e. its functions were independently initialized)
then a modular $\camech$ enforces dynamic modularity:
\begin{theorem}[\textbf{modular credit assignment}] \label{thm:modular_credit_assignment}
Dynamic modularity is enforced at learning iteration $\learnindex$ if and only if static modularity holds at iteration $i=0$ and the credit assignment mechanism satisfies the modularity constraint.
\end{theorem}
Initializing different functions with different weights is not sufficient to guarantee dynamic modularity.
The gradients produced by $\camech$ must be independent as well.
If $\camech$ were not modular it would be impossible for it to modify a function without simultaneously inducing a dependence with another, other than via non-generic instances where $\delta_\traceindex$ has a simple description, i.e. $\delta_\traceindex = 0$, which, unless imposed, are unlikely to hold over all iterations of learning.
\section{An Algorithmic Causal Model of Learning} \label{sec:an_algorithmic_causal_model_of_learning}
We can determine the dynamic modularity of a learning algorithm if we can evaluate the modularity constraint, 
but evaluating it is not practical in its current form because algorithmic information is generally incomputable.
This section proposes to bypass this incomputability by translating the constraint into a $d$-separation criterion on the causal structure of $\camech$, defined as part of one single causal graph of the learning process, which combines both the model of execution and the model of credit assignment.
The challenge to constructing this graph is that $\allfunc$ are treated as functions in $\mexec$ but as data in $\metamdp$, so it is not obvious how to reconcile the two in the same graph.
We solve this by treating the function application operation
\texttt{APPLY}~\citep{abelson1996structure}\footnote{This operation is known in $\lambda$-calculus as $\beta$-reduction.}, where $\forall \texttt{f},x, \; \texttt{APPLY}(\texttt{f},x) := \texttt{f}(x)$, as itself a function in a computational graph, enabling us to treat both \texttt{f} and $x$ as variables in the same flattened dynamic graph (Fig.~\ref{fig:main}c).

\begin{figure*}[!h]
    \centering
    \includegraphics[width=0.9\textwidth]{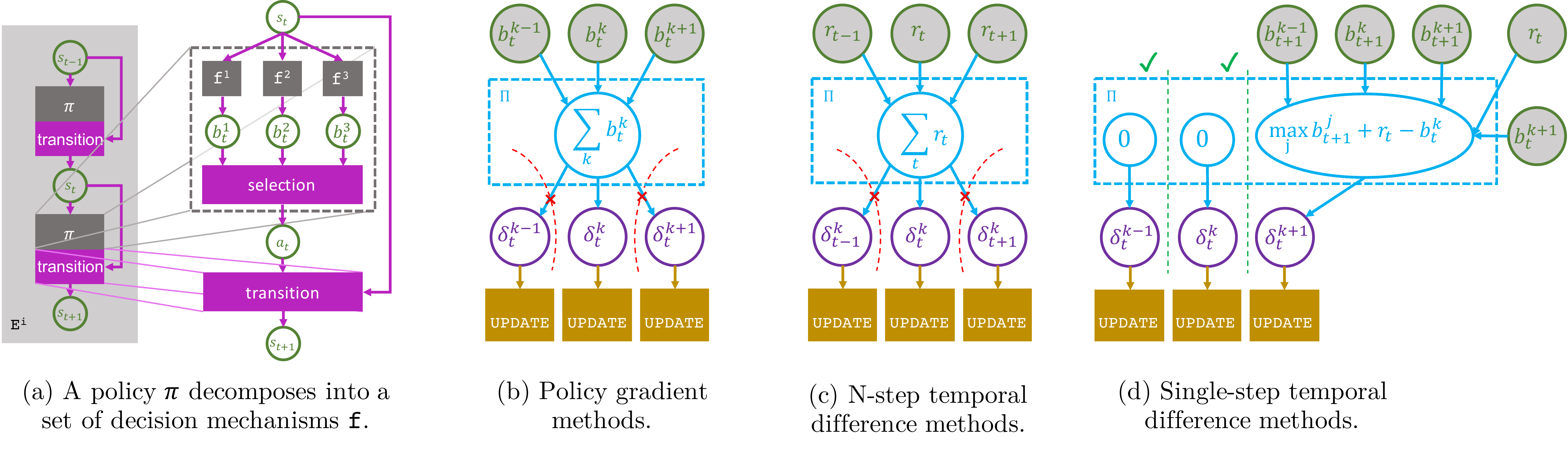}
    \caption{
    \small{\textbf{Modularity in RL.}
    In RL, the forward pass is a rollout in the MDP.
    (a) The societal decision-making framework exposes the learnable decision mechanisms of the policy as separate components in the model of execution.
    The bids $b$ represent either action probabilities are estimated action-specific $Q$-values.
    The credit assignment mechanisms of (b) policy gradient methods and (c) TD($n>1$) methods, like using Monte Carlo estimation, contain shared hidden variable and thus do not produce algorithmically  independent gradients $\delta$.
    (d) TD($0$) methods have modular credit assignment mechanisms in generic cases.
    The red crosses indicate a lack of $d$-separation, whereas the green checkmarks do.
    }}
    \label{fig:modularity_in_rl}
\end{figure*}

\begin{lemma}[\textbf{algorithmic causal model of learning}] \label{lem:algorithmic_causal_model_of_learning}
Given a model of execution $\mexec$ and of credit assignment $\metamdp$, define the \textbf{algorithmic causal model of learning} (\modelacronym) as a dynamic computational graph $\acl$ of the learning process.
We assume $\camech$ has its own internal causal structure with internal variable and function nodes.
The function nodes of $\acl$ are \texttt{APPLY}, \texttt{UPDATE}, and internal function nodes of $\camech$.
The variable nodes of $\acl$ are $x$, \texttt{f}, $\delta$, and internal variable nodes of $\camech$.
\texttt{APPLY} and \texttt{UPDATE} are assumed to have length $O(1)$.
The internal function nodes of $\camech$ jointly independent, and along with the variable nodes of $\acl$, are assumed to not have length $O(1)$.
Then these variable nodes satisfy the algorithmic causal Markov condition with respect to $\acl$ for all steps of credit assignment.
\end{lemma}
\modelacronym~is the bridge that brings tools from algorithmic causality~\citep{janzing2010causal} to bear on analyzing not simply the algorithmic independence of variables, but algorithmic independence of \emph{functions} in general learning algorithms.
The learnable mechanisms are no longer considered to have length $O(1)$ as is assumed in the model of execution.
With~\modelacronym, we define a criterion to test whether the modularity constraint holds by direct inspection:
\begin{theorem}[\textbf{modularity criterion}] \label{thm:modularity_criterion}
If $\acl$ is faithful, the modularity constraint holds if and only if for all $i$, outputs $\delta_\traceindex$ and $\delta_\othertraceindex$ of $\camech$ are $d$-separated by its inputs $\alldata$ and $\allfunc$.
\end{theorem}
We generally have access to the true computational graph, because the learning algorithm was programmed by us.
Thus Thm.~\ref{thm:modularity_criterion} enables us to evaluate, before any training, whether a learning algorithm satisfies the modularity constraint by simply inspecting $\acl$ for $d$-separation (Fig.~\ref{fig:main}d,e), giving us a practical tool to both design and evaluate learning algorithms on the basis of dynamic modularity.
\section{Modularity in Reinforcement Learning} \label{sec:modularity_in_reinforcement_learning}
We now apply the modularity criterion to evaluate the dynamic modularity of two major classes of RL algorithms~\citep{sutton2020reinforcement} -- action-value and policy-gradient methods.
The modularity criterion unlocks the use graphical language for our analysis, which simplifies the proofs.
We define a common model of execution for all algorithms within the~\societyacronym~framework from~\S\ref{sec:societal_decision_making} that enables us to compare the causal structures of their different credit assignment mechanisms under~\modelacronym.
We find that in the general function approximation setting, assuming acyclic decision sequences, the cloned Vickrey society (\clonedvickreysociety, \S\ref{sec:societal_decision_making}) is the only algorithm to our knowledge so far that produces reinforcement learners that exhibit dynamic modularity.

\subsection{From Monolithic Policies to Decision Mechanisms} \label{sec:from_monolithic_policies_to_decision_mechanisms}
As mentioned in~\S\ref{sec:societal_decision_making}, and as motivated by our taqueria example, we are interested in analyzing the independence of different decisions, so we need to adapt the model of execution we gave as an example for MDPs in~\S\ref{sec:learning_algorithms_are_dynamic_systems} to treat the functions that control each decision as separate mechanisms.

We observe from the~\societyacronym~framework that any discrete-action policy $\pi$ with $N$ actions can be decomposed into a set of mechanisms computing a ``bid'' $b^k_{s_\traceindex}$ for each \textbf{decision} $k$ (i.e., a value of the action variable, recall~\S\ref{sec:societal_decision_making}) at the given state $s_\traceindex$, and an independent selection mechanism that selects a decision given the bids (Fig.~\ref{fig:modularity_in_rl}a).
Define a \textbf{decision mechanism} as the function that computes a bid.
For policy-gradient methods, a bid corresponds to the action probability for a particular action $p(a=k | \cdot)$, and the selection mechanism is the stochastic sampler for a categorical variable.
For action-value methods, a bid corresponds to the estimated $Q$-value for a particular action, $Q(\cdot, a=k)$, and the selection mechanism could be an $\varepsilon$-greedy sampler or a Vickrey auction~\citep{chang2020decentralized}.
Often decision mechanisms share weights (e.g. DQN~\citep{mnih2015human}) and thus are algorithmically dependent, but for some algorithms they do not, as in~\clonedvickreysociety.
Then, by absorbing the transition function $\texttt{T}$ and reward function $\texttt{R}$ into \texttt{APPLY}, the function nodes $\allfunc$ of our model of execution are the decision mechanisms, which each take as input $s_\traceindex$, and produce as output the tuple $(b_{s_\traceindex}^\funcindex, s_{\traceindex+1}, r_\traceindex, w_\traceindex^\funcindex)$, where $w_\traceindex$ is a binary flag that indicates whether the selection chose its corresponding action.
The execution trace $\alldata$, which we call a \textbf{decision sequence}, records the values of these variables in a rollout.

\subsection{The Modularity of RL Algorithms} \label{sec:the_modularity_of_rl_algorithms}
We now ask which action-value and policy-gradient methods exhibit dynamic modularity by evaluating whether their credit assignment mechanisms satisfy the modularity criterion and whether their decision mechanisms share weights.

\paragraph{Which RL algorithms satisfy the modularity criterion?} \label{sec:which_rl_algorithms_satisfy_the_modularity_criterion}
The modularity criterion can be violated if there exists a shared hidden variable in the causal structure of $\camech$ that couples together the gradients $\delta$, which causes the $\delta^\funcindex_\traceindex$'s to not be $d$-separated given $\trace$ and $\allfunc$ (Fig.~\ref{fig:modularity_in_rl}b-d).

For all policy gradient methods, the gradient into the action probabilities includes a normalization term $\sum_k b^k$ as a shared hidden variable (Fig.~\ref{fig:modularity_in_rl}b):
\begin{corollary}[\textbf{policy gradient}] \label{cor:policy_gradient}
All policy gradient methods do not satisfy the modularity criterion.

\end{corollary}
We divide action-value methods into single-step and $n$-step (where $n > 1$) temporal difference methods, abbrv. TD($0$) and TD($n>1$) respectively.
For TD($n>1$) methods, such as those that use Monte Carlo (MC) estimation of returns, TD($\lambda$)~\citep{sutton1985temporal}, or generalized advantage estimation~\citep{schulman2015high}, this shared hidden variable is a sum of estimated returns or advantages at different steps of the decision sequence (Fig.~\ref{fig:modularity_in_rl}c):
\begin{corollary}[\textbf{n-step TD}] \label{cor:nsteptd}
All TD($n>1$) methods do not satisfy the modularity criterion.
\end{corollary}
This leaves only TD($0$) methods.
If the decision mechanism $\thisfunc$ were selected (i.e. $w_\traceindex^\funcindex = 1$) at step $i$, these methods produce, for some function $g$, gradients as $\delta^\funcindex_\traceindex: = g(b^\funcindex_{s_\traceindex}, s_\traceindex, s_{\traceindex+1}, r_\traceindex, \allfunc)$.
Otherwise, $\delta^\funcindex_\traceindex := 0$.
For example, for $Q$-learning, $g$ is the TD error $[\max_\otherfuncindex b^\otherfuncindex_{s_{\traceindex+1}} + r_\traceindex - b^\funcindex_{s_\traceindex}]$ (Fig.~\ref{fig:modularity_in_rl}d), where $[\max_\otherfuncindex b^\otherfuncindex_{s_{\traceindex+1}}]$ is computed from $s_{\traceindex+1}$ and $\allfunc$.
The only hidden variable is $[\max_\otherfuncindex b^\otherfuncindex_{s_{\traceindex+1}}]$.
It is only shared when the decision sequence $\trace$ contains a cycle where two states $s_\traceindex$ and $s_\traceindex'$ transition into the same state $s_{\traceindex+1}$.
In this cyclic case, the credit assignment mechanism 
would not satisfy the modularity criterion.
Otherwise it does:
\begin{corollary}[\textbf{single-step TD}] \label{cor:td0}
TD($0$) methods satisfy the modularity criterion for acyclic $\trace$.
\end{corollary}
As cyclic $\trace$ are non-generic cases that arise from specific settings of $\trace$, we henceforth restrict our analysis to the acyclic case, justifying this restriction similarly to the justification of assuming faithfulness in other causal literature.

\paragraph{Which RL algorithms exhibit dynamic modularity?}
We have identified TD($0$) methods as the class of RL algorithms that satisfy the modularity criterion.
By Thm.~\ref{thm:modular_credit_assignment}, 
whether they satisfy dynamic modularity now depends on whether they satisfied static modularity at initialization ($i=0$).
We assume random initialization of $\allfunc$, so the only source of dependence among $\allfunc$ is if they share parameters.

In the tabular setting, decision mechanisms are columns of the $Q$-table corresponding to each action.
Because these columns do not share parameters, $Q$-learning~\citep{watkins1992q}, SARSA~\citep{rummery1994line}, and \clonedvickreysociety~exhibit dynamic modularity:
\addtocounter{theorem}{-3}
\setcounter{corollary}{0}
\begin{corollary}[\textbf{tabular}] \label{cor:tabular}
In the tabular setting, Thm.~\ref{thm:modular_credit_assignment} holds for $Q$-learning, SARSA, and \clonedvickreysociety.
\end{corollary}
In the general function approximation setting, static modularity requires decision mechanisms to not share weights, which eliminates DQN~\citep{mnih2015human} and its variants.
\begin{corollary}[\textbf{function approximation}] \label{cor:funcapprox}
In the function approximation setting, Thm.~\ref{thm:modular_credit_assignment} holds for TD($0$) methods whose decision mechanisms do not share parameters.
\end{corollary}
\addtocounter{theorem}{3}
To our knowledge,~\clonedvickreysociety~is the only proposed TD($0$) method with this property, but it is straightforward to make existing TD($0$) methods exhibit dynamic modularity by using separate networks for estimating the $Q$-value of each decision.

\paragraph{Summary.}
If we want dynamic modularity, then we need the decision mechanisms to not share parameters and the credit assignment mechanism to not contain a shared hidden variable that induced algorithmic dependence among the gradients it outputs.
An RL algorithm with dynamic modularity makes it possible for individual decision mechanisms to be modified independently without an accompanying modification to other decision mechanisms.
\definecolor{actionB}{RGB}{247, 101, 9}
\definecolor{actionC}{RGB}{240, 182, 0}
\definecolor{actionD}{RGB}{131, 189, 69}

\begin{figure}
    \centering
    \includegraphics[width=0.45\textwidth]{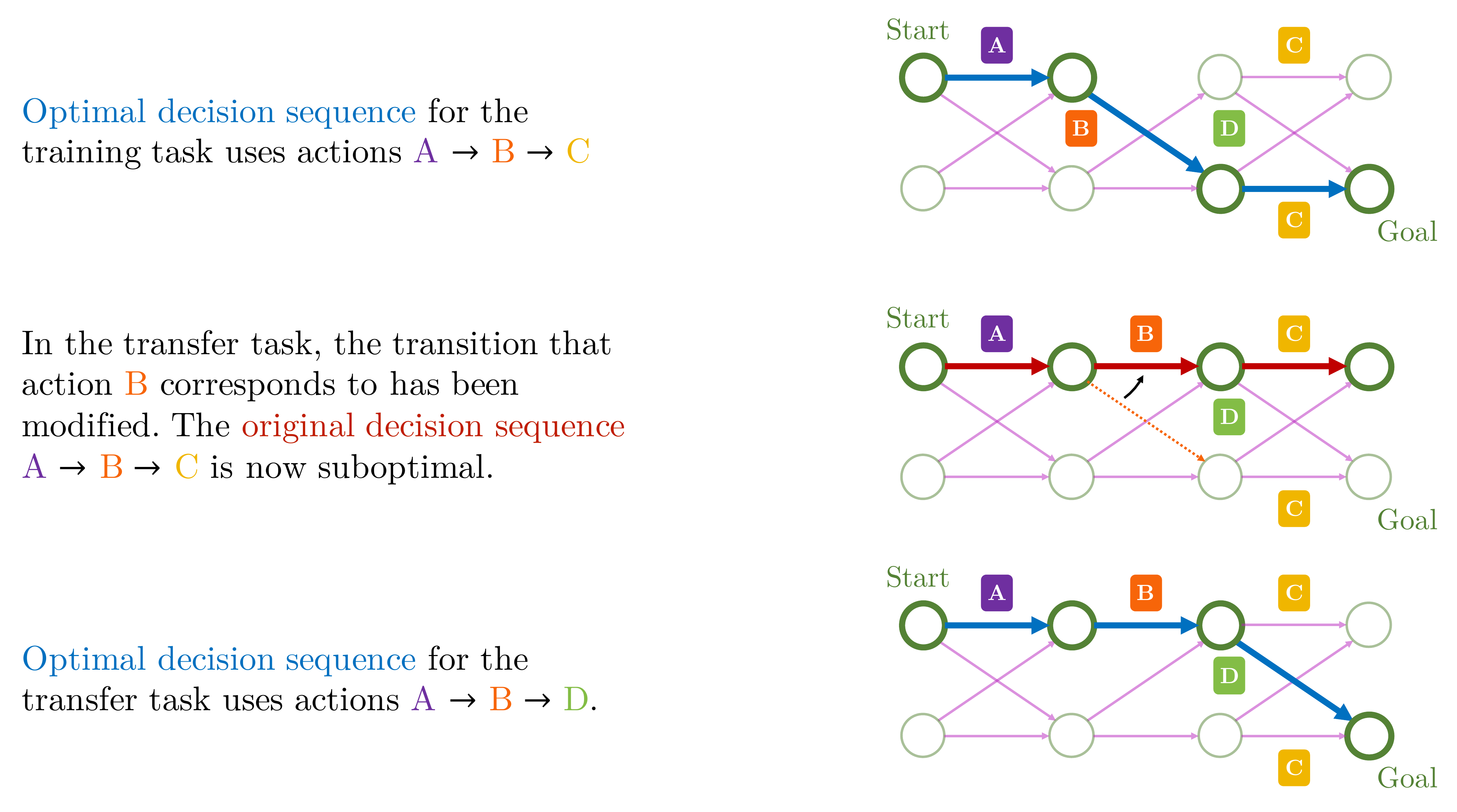}
    \caption{\small{
    \textbf{How transfer tasks are generated.}
    We consider transfer problems where the optimal decision sequence of the transfer task differs from that of the training task by a single decision.
    As above, the transfer MDP and the training MDP differ in that the effect of action \textcolor{actionB}{$\mathbf{B}$}; all other transitions remain the same.
    The agent must learn to choose action \textcolor{actionD}{$\mathbf{D}$} instead of \textcolor{actionC}{$\mathbf{C}$} while re-using other previously optimal decisions.
    }}
    \label{fig:intervention_on_the_transition_function}
\end{figure}

\section{Simple Experiments} \label{sec:experiments}
This paper is motivated by the hypothesis that modularity enables flexible adaptation.
To test this hypothesis requires (1) a method for determining whether a learning algorithm is modular and (2) a metric for evaluating flexible adaptation.
The previous sections have contributed (1).
The metric we use for (2) is the comparative transfer efficiency of an algorithm that exhibits dynamic modularity with respect to one that does not.
We consider transfer problems that require modifying only one decision in a previously optimal decision sequence needs to be changed, similar to our motivating example with Gusteau's taqueria (\S\ref{sec:introduction}).

Our evaluation focuses on discrete-action on-policy RL algorithms since many factors that influence the learning of off-policy methods are still not well understood~\citep{achiam2019towards,kumar2020implicit,van2018deep,fu2019diagnosing}.
Specifically we compare three algorithms that span the spectrum of action-value and policy-gradient methods.
\clonedvickreysociety~represents a method that exhibits dynamic modularity.
PPO~\citep{schulman2017proximal} represents a method that is not modular at all.
PPOF is a modification of PPO whose where each action logit is computed by a different network, and represents a method that exhibits static modularity at initialization but not dynamic modularity during learning.

We designed our experiments to be as minimal as possible to remove confounders.
States are represented as binary vectors.
The reward is given at the end of the episode and is 1 if the task is solved and 0 otherwise.
The relationship between the training and transfer MDP is given by an intervention in the MDP transition function (Fig.~\ref{fig:intervention_on_the_transition_function}).

\subsection{An Enumeration of Transfer Problems} \label{sec:an_enumeration_of_transfer_problems}
\begin{figure*}
    \centering
    \includegraphics[width=\textwidth]{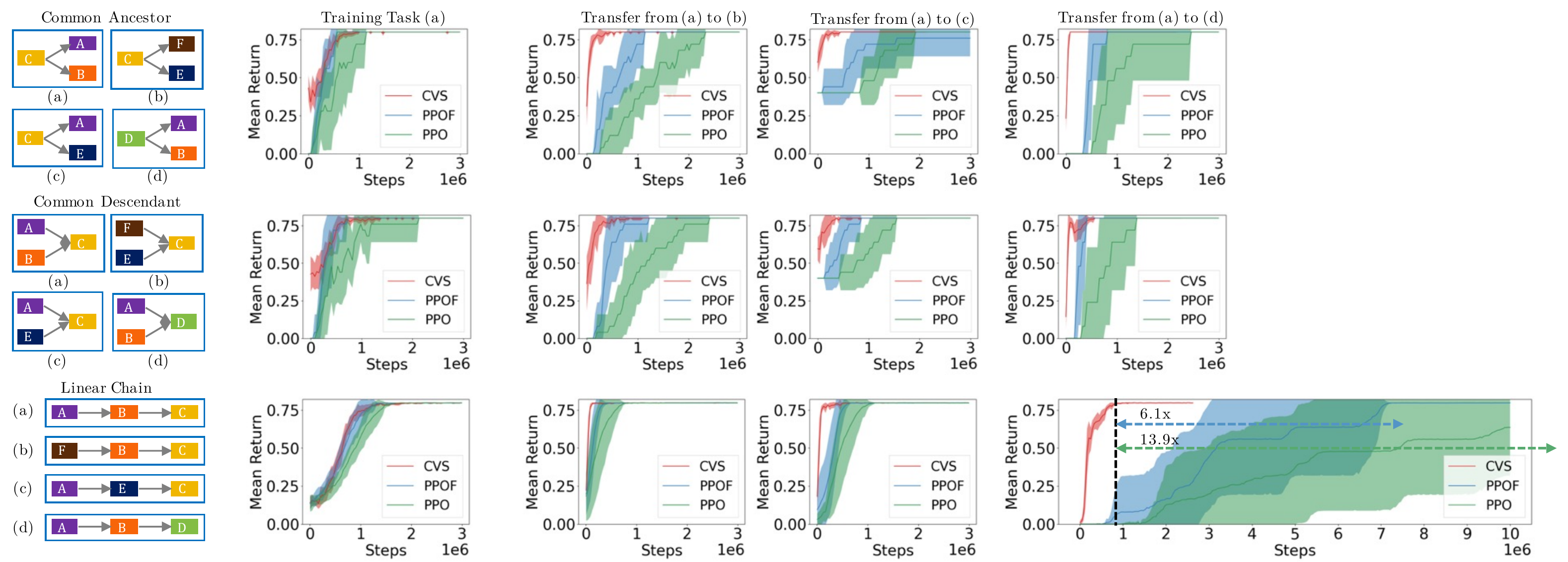}
    \caption{\small{\textbf{Transfer problems involving triplets of decisions.}
    For each task topology (leftmost column) we have a training task, labeled (a) and three independent transfer tasks, labeled (b,c,d).
    Each transfer task is a different way to modify the training task's MDP.
    \clonedvickreysociety~consistently exhibits higher sample efficiency than both PPO and PPOF showing that dynamic modularity correlates with more efficient transfer.
    Notably the gap between~\clonedvickreysociety~and the other methods in the bottom-right (e.g. 13.9x more efficient than PPO) is so wide that we had to extend the chart width.
    We set the convergence time as the first time after which the return deviates by no more than $\varepsilon = 0.01$ from the optimal return, 0.8, for 30 epochs of training.
    Shown are runs across ten seeds.
    }}
    \label{fig:task_graphs}
\end{figure*}

Similar to how analysis of $d$-separation is conducted with triplets of nodes, we enumerated all possible topologies of triplets of decisions: \textit{linear chain}, \textit{common ancestor}, and \textit{common descendant} (Fig.~\ref{fig:task_graphs}, left column).
For each topology we enumerated all ways of making an isolated change to an optimal decision sequence.
The \textit{common ancestor} and \textit{common descendant} topologies involve multi-task training for two decision sequences of length two, while \textit{linear chain} involves single-task training for one decision sequence of length three.
For example, in Fig.~\ref{fig:task_graphs}, the optimal decision sequence for the \textit{linear chain} training task is $A \rightarrow B \rightarrow C$.
For each topology we have a training task and three independent transfer tasks.
Each transfer task represents a different way to modify the MDP of the training task.
This single comprehensive task suite (Fig.~\ref{fig:task_graphs}) enables us to ask a wide range of questions.
The answers to the questions that follow are scoped only to our stated experimental setup.

\begin{figure}
    \centering
    \includegraphics[width=0.40\textwidth]{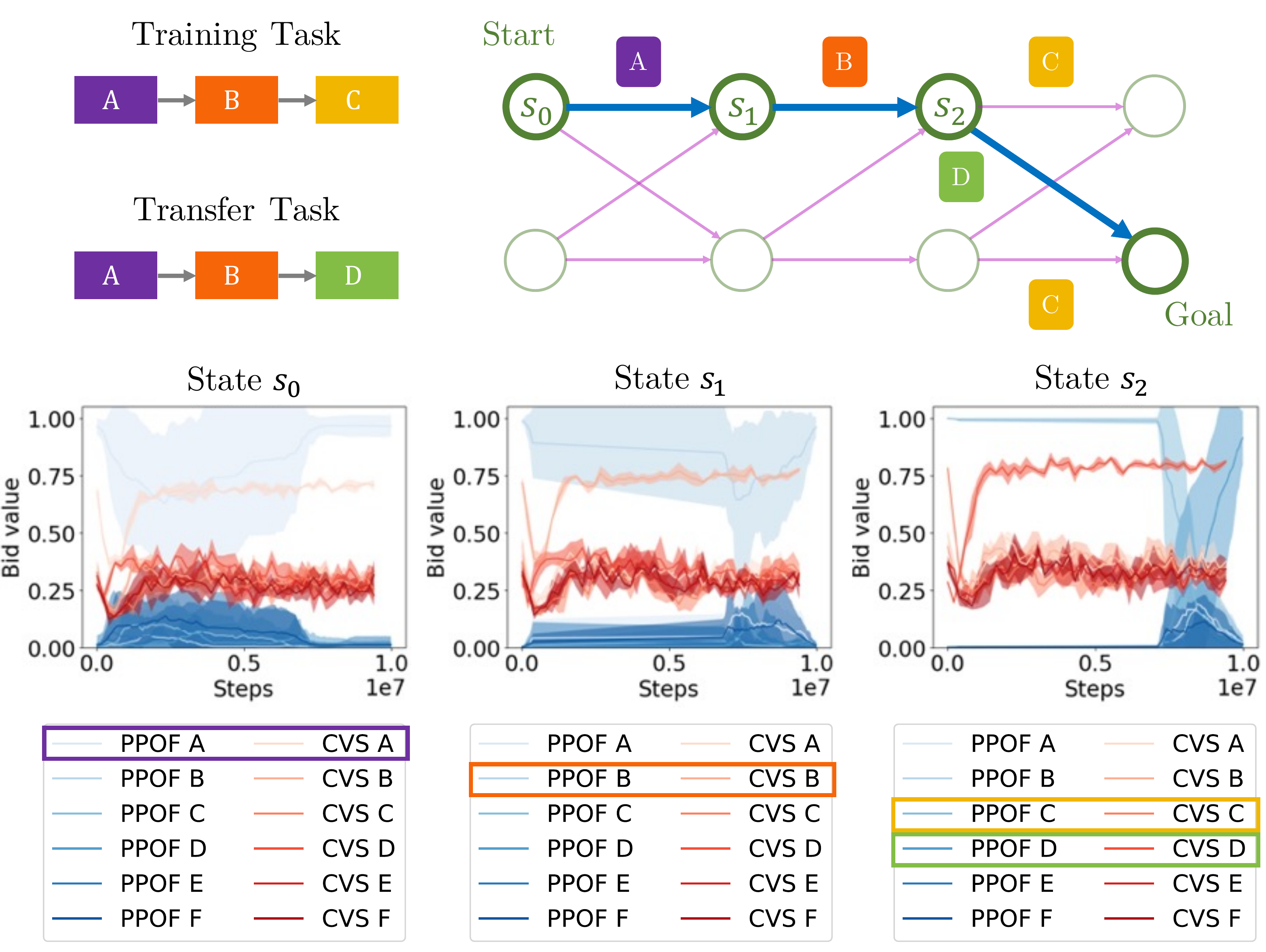}
    \caption{\small{\textbf{How the decision mechanisms change during transfer.}
    Shown the three states of the decision sequence.
    The optimal last decision must change from action C (purple) to action D (green).
    \clonedvickreysociety~modifies its bids independently.
    The bids for PPOF are coupled together across decision mechanisms and across time.
    }}
    \label{fig:bidding_curves}
\end{figure}

\paragraph{Does dynamic modularity improve transfer efficiency?}
Yes, at least in these experiments.
For each of the nine transfer settings (rightmost three columns) in Fig.~\ref{fig:task_graphs},~\clonedvickreysociety~(red) transfers consistently more efficiently than both PPO (green) and PPOF (blue), despite having comparable training efficiency in the training task (second column from left).
The variance among the different runs is also lower for~\clonedvickreysociety.

\paragraph{How does where a decision needs to be modified in the decision sequence affect transfer efficiency?}
The improvement in transfer efficiency is especially pronounced in the trend shown in the bottom row of Fig.~\ref{fig:task_graphs} for \emph{linear chain}.
The later the decision that needs to be modified appears in the decision sequence, the wider the gap between~\clonedvickreysociety~and the other two methods, to the point that we had to widen the plot width.
Our theory (Thm.~\ref{thm:modular_credit_assignment}) offers one possible explanation.
Considering the bottom-right plot of Fig.~\ref{fig:task_graphs}, the transfer task requires modifying the last decision and keeping the previous two the same.
But the lack of independent gradients and parameters in PPO and PPOF seems to have affected correct decision mechanisms in the first two steps based on the errors encountered by the decision mechanism in the last step, seemingly causing the previous decision mechanisms to ``unlearn'' originally optimal behavior, then relearn the correct behavior again, as shown in the plots for ``state $s_0$'' and ``state $s_1$'' in Fig.~\ref{fig:bidding_curves} for PPOF.
This slow unlearning and relearning seems to be a reason for the lower transfer efficiency of PPO and PPOF.
It is as if Colette in Gusteau's taqueria (\S\ref{sec:introduction}) stopped heating tortillas because of the angry reviews about meat contamination but then realized that she should still be heating tortillas after all.

\paragraph{Does dynamic modularity enable independent modification of decision mechanisms in practice?}
While theory tells us that decision mechanisms can be modified independently within a single credit assignment update, in practice transfer learning requires multiple credit assignment updates to converge.
Across multiple credit assignment updates, the decision mechanisms would no longer be independent, even for algorithms that exhibit dynamic modularity, but it is also expected that the functions of a learner should learn to work together over the course of learning in any case.
Nonetheless, Fig.~\ref{fig:bidding_curves} shows that the lack of a softmax tying the bids of~\clonedvickreysociety~together enables them to change more independently and rapidly than PPOF.

\paragraph{How much of transfer efficiency is due to modular credit assignment than network factorization?}
This question pits our theory against a competing explanation: that network factorization alone (represented by PPOF) is responsible for improved transfer efficiency.
Though PPOF is more efficient than PPO in training and transfer, PPOF is consistently less efficient than~\clonedvickreysociety~in transfer while being similarly efficient in training.
This suggests that network factorization is not a sufficient explanation, leaving our theory of dynamic modularity still standing.

\subsection{Modularity and Forgetting}
\begin{figure*}
    \centering
    \includegraphics[width=\textwidth]{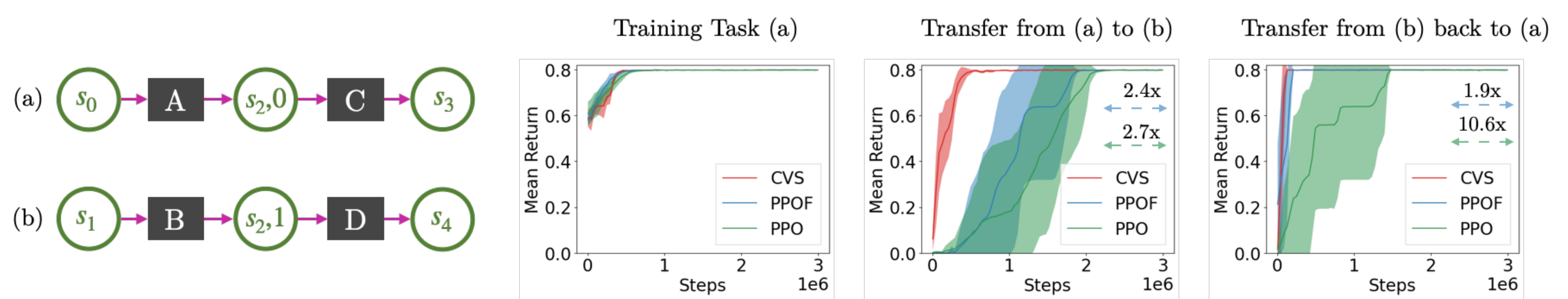}
    \caption{\small{\textbf{Modularity and forgetting.}
    The optimal solutions for tasks (a) and (b) involve a disjoint set of decisions: $A\rightarrow C$ for task (a) and $B \rightarrow D$ for task (b).
    We first train on task (a), then transfer from (a) to (b), then transfer back from (b) to (a).
    The purpose of this experiment is to test whether dynamic modularity improve the agent's ability to preserve optimal behavior on a previous task after having trained to convergence on a different task in a different context.
    While both~\clonedvickreysociety~and PPO have similar sample efficiency when initially training on task (a),~\clonedvickreysociety~ is more than ten times more sample efficient than PPO when transferring back from (b) to (a), suggesting that PPO ``forgot'' the optimal behavior for task (a) when training on task (b), which is not the kind of forgetting we want in learning agents.
    }}
    \label{fig:invariance}
\end{figure*}

A desirable consequence of having the capacity to independently modify learnable mechanisms is the ability to \emph{not} modify mechanisms that need not be modified: we would not want the agent to forget optimal behavior in one context when it trains on a different task in a different context.
We now test whether dynamic modularity contributes to this ability.
The experimental setup is shown in Fig.~\ref{fig:invariance}.
There are four possible values for the action, $A, B, C, D$.
In task (a), the optimal decision sequence is $A \rightarrow C$, starting at state $s_0$ and passing through state $s_2$, which has a context bit flipped to $0$.
In task (b), the optimal decision sequence is $B \rightarrow D$, starting at state $s_1$ and passing through state $s_2$, which has a context bit flipped to $1$.
Though the optimal states for task (a) are disjoint from the optimal states for task (b), the decision mechanisms corresponding to $A, B, C, D$ are present for both tasks.
We first train on task (a), then transfer from (a) to (b), then transfer back from (b) to (a).

\paragraph{Does dynamic modularity improve the agent's ability to preserve optimal behavior on a previous task after having trained to convergence on a different task?}
To test this, we compare~\clonedvickreysociety~and PPO's sample efficiency when transferring back from (b) to (a).
Fig.~\ref{fig:invariance} shows that even when both~\clonedvickreysociety~and PPO have similar sample efficiency when initially training on task (a),~\clonedvickreysociety~ is more than ten times more sample efficient than PPO when transferring back from (b) to (a).
Our explanation for this phenomenon is that the lack of algorithmic independence in the decision mechanisms of PPO causes the decision mechanisms for actions $A$ and $C$ to be significantly modified when PPO transfers from (a) to (b), even when these actions do not even participate in the optimal decision sequence for task (b).
The low sample efficiency when transferring back from (b) to (a) suggests that PPO ``forgot'' the optimal behavior for task (a) when training on task (b), which is not the kind of forgetting we want in flexibly adaptable agents.

\paragraph{How much of this ability to preserve previously optimal behavior due to modular credit assignment than network factorization?}
PPOF is similarly inefficient as PPO compared to~\clonedvickreysociety~in transferring from (a) to (b), which is consistent with our findings from~\S\ref{sec:an_enumeration_of_transfer_problems}.
Interestingly, PPOF seems to be just as efficient at transferring back from (b) to (a) as~\clonedvickreysociety, which seems to suggest that the primary cause for the forgetfulness of PPO, at least in this experiment, is less due to lack of independent gradients but more to lack of network factorization.
This experiment suggests a need for an explanatory theory to identify under which circumstances independent gradients are more influential to flexible adaptation than network factorization, and vice versa, as well as a means for quantifying the degree of influence each has.

\section{Discussion} \label{sec:discussion}
The hypothesis that modularity can enable flexible adaptation requires a method for determining whether a learning system is modular.
This paper has contributed the modularity criterion (Thm.~\ref{thm:modularity_criterion}) as such a method.

The consistency of how dynamic modularity in on-policy reinforcement learning correlates with higher transfer efficiency in our experiments suggests a need for future work to provide an explanatory theory for exactly how dynamic modularity contributes to flexible adaptation as well as to test whether the same phenomenon can be observed with other classes of learning algorithms, other transfer problems, and other domains.
The modularity criterion is a binary criterion on algorithmic independence or lack thereof, but our experiments also suggest a need for future work in quantifying algorithmic causal influence if we want to relax the criterion to a softer penalty on algorithmic mutual information.
Moreover, our work highlights a need for a formalism that allows for meaningful statements about independence among concrete strings within a specific Turing machine, rather than statements that only hold asymptotically.

Learning algorithms are simply only one example of the more general concept of the dynamic computational graph introduced in this paper, which we have shown with Lemma~\ref{lem:algorithmic_causal_model_of_learning} can be used to analyze the algorithmic independence of functions that evolve over time.
The connection we have established among credit assignment, modularity, and algorithmic information theory, in particular the link between learning algorithms and algorithmic causality, opens many opportunities for future work, such as new ways of formalizing inductive bias in the algorithmic causal structure of learning systems \emph{and} the learning algorithms that modify them.
\section*{Acknowledgements}
We thank Kumar Krishna Agrawal, Pim de Haan, Daniel Filan, Abhishek Gupta, Michael Janner, Giambattista Parascandolo, Sam Toyer, Olivia Watkins, Marvin Zhang for feedback on earlier drafts of the paper.
We thank Dominik Janzing, Aviral Kumar, and Jason Peng for useful discussions.
This research was supported by the AFOSR grant FA9550-18-1-0077, and computing resources from Amazon Web Services and Microsoft Azure.
MC is supported by the National Science Foundation Graduate Research Fellowship Program.
\bibliography{bib}
\bibliographystyle{icml2021}
\clearpage
\onecolumn
\begin{appendices}

\section{Background} \label{appdx:background}
This section supplements \S\ref{sec:algorithmic_causality} with more background on algorithmic information theory and standard causality.
For a more thorough treatment on the foundational mathematics and formalism, please refer to~\citet{li2008introduction} for algorithmic information theory, to~\citet{pearl2009causality} for standard causality, and  to~\citet{janzing2010causal,peters2017elements} for algorithmic causality.

\subsection{Notation} \label{appdx:notation_and_terminology}
We denote with bolded uppercase monospace a computation graph at a single level of abstraction (e.g. the model of execution $\compgraph$, the model of credit assignment $\metamdp$).
We denote with blackboard bold (e.g. the algorithmic causal model of learning $\mathbb{L}$) a computation graph that represents multiple levels of abstraction.

We denote binary strings that represent the data nodes in $\mathbb{L}$ with lower case (e.g. $x$ or $\texttt{f}$), where script ($x$) is used to emphasize that the string typically represents a variable and monospace ($\texttt{f}$) is used to emphasize that the string typically represents a function.
We use bolded lower case (e.g. $\trace$, $\bm{\delta}$, $\allfunc$) to indicate a group of binary strings.
We denote the function nodes in $\mathbb{L}$ (e.g., $\texttt{APPLY}$, $\texttt{UPDATE}$) with uppercase.

We write $\texttt{f}(x) \rightarrow y$ to mean ``a program $\texttt{f}$ that takes a string $x$ as input and produces a string $y$ as output.''

\subsection{Background on algorithmic causality} \label{appdx:algorithmic_causality}
The formalism of algorithmic causality derives from~\citet{janzing2010causal,peters2017elements}, which builds upon algorithmic statistics~\citep{gacs2001algorithmic}.
Here we directly restate or paraphrase additional relevant definitions, postulates, and theorems from~\citet{janzing2010causal} and~\citet{gacs2001algorithmic}.

\subsubsection{Algorithmic information theory} \label{appdx:algorithmic_information_theory}
\paragraph{Kolmogorov complexity}
Kolmogorov complexity~\citep{solomonoff1960preliminary,solomonoff1964formal,kolmogorov1965three,chaitin1966length,chaitin1975theory,li2008introduction} is a function $K: \{0, 1\}^* \rightarrow \mathbb{N}$ from the binary strings $\{0, 1\}^*$ to the natural numbers $\mathbb{N}$ that represents the amount of information contained in an object (represented by a binary string).
\begin{definition}[\textbf{Kolmogorov-complexity}] \label{def:kolmogorov-complexity}
Given a universal Turing machine and universal programming language as reference, the \textbf{Kolmogorov complexity} $K(s)$ is the length of the shortest program that generates $s$.
The \textbf{conditional Kolmogorov complexity} $K(y \given x)$ of a string $y$ given another string $x$ is the length of the shortest program that generates $y$ given $x$ as input.
Let the shortest program for string $x$ be denoted as $x^*$.
The \textbf{joint Kolomogorov complexity} $K(x, y)$ is defined as:
\begin{equation*}
    K\left(x, y\right) \overset{+}{=} K\left(x\right) + K\left(x \mgiven y^*\right) \overset{+}{=} K\left(y\right) + K\left(y \mgiven x^*\right).
\end{equation*}
\end{definition}
The \textbf{invariance theorem}~\citep{kolmogorov1965three} states that the Kolmogorov complexities of two strings written in two different universal languages differ only up to an additive constant.
Therefore, we can assume any reference universal language for defining $K$ (e.g. Python) and work with equalities ($\overset{+}{=}$) and inequalities ($\overset{+}{\geq}$, $\overset{+}{\leq}$) up to an additive constant.

\paragraph{Algorithmic mutual information}
Algorithmic mutual information $I$ measures the amount of information two objects have in common:
\begin{definition}[\textbf{algorithmic mutual information}] \label{def:algorithmic_mutual_information}
The \textbf{algorithmic mutual information} of two binary strings $x,y$ is 
\begin{equation*}
    I\left(x:y\right) \overset{+}{=} K\left(x\right) + K\left(y\right) - K\left(x,y\right).
\end{equation*}
The \textbf{conditional algorithmic mutual information} of strings $x,y$ given string $z$ is
\begin{equation*}
    I\left( x:y \mgiven z \right) \overset{+}{=} K\left(x \mgiven z \right) + K\left(y \mgiven z \right) - K\left(x,y \mgiven z \right)
\end{equation*}
\end{definition}
We can intuitively think of $I(x : y \given z)$ as, given $z$, the number of bits that can be saved when describing $y$ knowing the shortest program that generated $x$.
We analogously extend this definition to the joint conditional algorithmic mutual information of strings $x_1, ..., x_n$ given strings $y_1, ..., y_m$, using ``...'' instead of ``:'' for notational a convenience:
\begin{definition}[\textbf{joint conditional algorithmic mutual information}]
\label{def:joint_conditional_algorithmic_mutual_information}
Given strings $x_1, ..., x_n$ and $y_1, ..., y_m$, the \textbf{joint algorithmic mutual information} of $x_1, ..., x_n$ given $y_1, ..., y_m$ is:
\begin{equation*}
    I\left( x_1, ..., x_n \mgiven y_1, ..., y_m \right) \overset{+}{=} \sum_{i=1}^n K\left(x_i \mgiven y_1, ..., y_m \right) - K\left(x_1, ..., x_n \mgiven y_1, ..., y_m\right).
\end{equation*}
\end{definition}

Then algorithmic independence is the property of two strings that says that the description of one cannot be further compressed given knowledge of the other.
\begin{definition}[\textbf{algorithmic conditional independence}] \label{def:algorithmic_conditional_independence}
Given three strings $x, y, z$, $x$ is \textbf{algorithmically conditionally independent} of $y$ given $z$, denoted by $x \indep y \given z$, if the additional knowledge of $y$ does not allow for stronger compression of $x$, given $z$.
That is:
\begin{equation*}
    x \indep y \given z \; \Leftrightarrow \; I\left(x : y \given z\right) \overset{+}{=} 0.
\end{equation*}
\end{definition}
Joint conditional independence of strings $x_1, ..., x_n$ given $y_1, ..., y_m$ is defined analogously as $I\left(x_1, ..., x_n \mgiven y_1, ..., y_m\right) \overset{+}{=} 0$.
\begin{lemma}[\textbf{algorithmic joint conditional independence}] \label{lemma:algorithmic_joint_conditional_independence}
If strings $x_1, ..., x_n$ are algorithmically jointly conditionally independent given strings $y_1, ..., y_m$, then
\begin{equation} \label{eqn:algorithmic_joint_zero_mutual_information}
    K\left(x_1, ..., x_n \mgiven y_1, ..., y_m\right) \overset{+}{=} \sum_{i=1}^n K\left(x_i \mgiven y_1, ..., y_m\right),
\end{equation}
\end{lemma}
meaning that, conditioned on knowing $y_1, ..., y_m$, the length of the joint description of $x_1, ..., x_n$ cannot be further compressed than sum of the lengths of the descriptions of the individual strings $x_i$.
The proof is by starting with $I\left(x_1, ..., x_n \mgiven y_1, ..., y_m\right) \overset{+}{=} 0$ and rearranging Def.~\ref{def:joint_conditional_algorithmic_mutual_information}.

We state as a lemma the following result from~\citet[Corollary $\Pi.8$]{gacs2001algorithmic} that states the mutual information of strings $x$ and $y$ cannot be increased by separately processing by functions $f$ and $g$.
\begin{lemma}[\textbf{information non-increase}] \label{lemma:information_nonincrease}
Let $f$ and $g$ be computable programs.
Then 
\begin{align}
    I\left(\texttt{f}(x) : \texttt{g}(y)\right) \overset{+}{\leq} I\left(x : y\right) + K(\texttt{f}) + K(\texttt{g}).
\end{align}
\end{lemma}
This intuitively makes sense: if $K(\texttt{f})$ is constant with respect to $x$ and $K(\texttt{g})$ is constant with respect to $y$ (i.e. $K(\texttt{f}) \overset{+}{=} 0$ and $K(\texttt{g}) \overset{+}{=} 0$), then mutual information cannot increase between $x$ and $y$ separately with $\texttt{f}$ and $\texttt{g}$.
In particular, if $x$ and $y$ were independent to begin with (i.e. $I(x : y) \overset{+}{=} 0$), then $I\left(\texttt{f}(x) : \texttt{g}(y)\right) \overset{+}{=} 0$.

\paragraph{Terminology}
In this paper, we regard the following statements about a program $\texttt{f}(x) \rightarrow y$ as equivalent:
\begin{itemize}
    \item ``$K(\texttt{f}) \overset{+}{=} 0$.''
    \item ``$\texttt{f}$ is an $O(1)$-length program.''
    \item ``$K(\texttt{f})$ is constant with respect to $x$.''
\end{itemize}
Note that $K(\texttt{f}) \overset{+}{=} 0$ implies that $\texttt{f}$ and $x$ are algorithmically independent (i.e. $I(\texttt{f} : x ) \overset{+}{=} 0$) because 
\begin{align*}
    0 \overset{+}{\leq} I\left(x : \texttt{f}\right) \overset{+}{=}  K\left(\texttt{f}\right) - K\left(\texttt{f} \mgiven x^*\right) \overset{+}{\leq} K\left(\texttt{f}\right) \overset{+}{=} 0.
\end{align*}

\subsubsection{Causality} \label{appdx:causality}
Before we review algorithmic causality, we first review some key concepts in standard causality: structural causal models and $d$-separation.

The following definition defines standard causal models over random variables as Bayesian networks represented as directed acyclic graphs (DAG), analogous to the algorithmic version we presented in Def.~\ref{def:computational_graph}.
\begin{definition}[\textbf{structural-causal-model}]
A \textbf{structural causal model} (SCM)~\citep{pearl1995causal,pearl2009causality} represents the assignment of random variable $X$ as the output of a function, denoted by lowercase monospace (e.g. \texttt{f}), that takes as input an independent noise variable $N_X$ and the random variables $\{PA_X\}$ that represent the parents of $X$ in a DAG:
\begin{equation}
    X := \texttt{f}(\{PA_X\}, N_X).
\end{equation}
Given the noise distributions $\mathbb{P}(N_X)$ for all variables $X$ in the DAG, SCM entails a joint distribution $\mathbb{P}$ over all the variables in the DAG~\citep{peters2017elements}.
\end{definition}

The graph-theoretic concept of $d$-separation is used for determining conditional independencies induced by a directed acyclic graph (see point 3 in Thm.~\ref{thm:equivalence_of_algorithmic_markov_conditions}):
\begin{definition}[\textbf{$d$-separation}] \label{def:d-separation}
A path $p$ in a $DAG$ is said to be d-separated (or blocked) by a set of nodes $Z$ if and only if
\begin{enumerate}
    \item $p$ contains a chain $i \rightarrow m \rightarrow j$ or fork $i \leftarrow m \rightarrow j$ such that the middle node $m$ is in $Z$, or 
    \item $p$ contains an inverted fork (or collider) $i \rightarrow m \leftarrow j$ such that the middle node $m$ is not in $Z$ and such that no descendant of $m$ is in $Z$.
\end{enumerate}
A set of nodes $Z$ \textbf{$d$-separates} a set of nodes $X$ from a set of nodes $Y$ if and only if $Z$ blocks every (possibly undirected) path from a node in $X$ to a node in $Y$.
\end{definition}

\subsubsection{Algorithmic causality}
For convenience, we re-state the technical content from \S\ref{sec:algorithmic_causality} here.

\textbf{Definition~\ref{def:computational_graph} (computational graph).}
\textit{Define a \textbf{computational graph} $\compgraph = (\alldata, \allfunc)$ as a directed acyclic factor graph (DAG) of variable nodes $\alldata = x_1, ..., x_N$ and function nodes $\allfunc = \texttt{f}^1, ..., \texttt{f}^N$.
Let each $x_j$ be computed by a program \texttt{f}$^j$ with length $O(1)$ from its parents $\{pa_j\}$ and an auxiliary input $n_j$.
Assume the $n_j$ are jointly independent: $n_j \indep \{n_{\neq j}\}$.
Formally, $x_j := \texttt{f}^j(\{pa_j\}, n_j)$, meaning that the Turing machine computes $x_j$ from the input $\{pa_j\}, n_j$ using the additional program $\texttt{f}^j$ and halts.}
\noindent

\textbf{Theorem~\ref{thm:algorithmic_causal_markov_condition} (algorithmic causal Markov condition).}
\textit{Let $\{pa_j\}$ and $\{nd_j\}$ respectively represent concatenation of the parents and non-descendants (except itself) of $x_j$ in a computational graph.
Then $\forall x_j$, $x_j \indep \{nd_j\} \given \{pa_j\}^*.$}
\noindent

\textbf{Postulate~\ref{post:faithfulness} (faithfulness).}
\textit{Given sets $S$, $T$, $R$ of nodes in a computational graph, $I(S : T | R^*) \overset{+}{=} 0$ implies $R$ d-separates $S$ and $T$.}
\noindent

The following theorem~\citet[Thm. 3]{janzing2010causal} establishes the connection between the graph-theoretic concept of $d$-separation with condition algorithmic independence of the nodes of the graph.

\begin{theorem}[\textbf{equivalence of algorithmic Markov conditions}] \label{thm:equivalence_of_algorithmic_markov_conditions}
Given the strings $x_1, ..., x_n$ and a computational graph,
the following conditions are equivalent:
\begin{enumerate}
    \item \textbf{Recursive form:} the joint complexity is given by the sum of complexities of each node $x_j$, given the optimal compression of its parents $\{pa_j\}$:
        \begin{equation*}
            K(x_1, ..., x_n) \overset{+}{=} \sum_{j=1}^n K(x_j | \{pa_j\}^*).
        \end{equation*}
    \item \textbf{Local Markov Condition:} Every node $x_j$ is independent of its non-descendants $\{nd_j\}$, given the optimal compression of its parents $\{pa_j\}$:
        \begin{equation*}
            I(x_j : nd_j | \{pa_j\}^*) \overset{+}{=} 0.
        \end{equation*}
    \item \textbf{Global Markov Condition:} Given three sets $S$, $T$, $R$ of nodes
        \begin{equation*}
            I(S : T | R^*) \overset{+}{=} 0 
        \end{equation*}
        if $R$ d-separates $S$ and $T$.
\end{enumerate}
\end{theorem}
Together, Thm.~\ref{thm:algorithmic_causal_markov_condition}, Post.~\ref{post:faithfulness}, and Thm.~\ref{thm:equivalence_of_algorithmic_markov_conditions} imply that variable nodes in a computational graph are algorithmically independent if and only if they are $d$-separated in the computational graph.

\section{Assumptions} \label{appdx:assumptions}
This section states our assumptions for the results that we prove in \S\ref{appdx:proofs}.
This paper analyzes learning algorithms from the perspective of algorithmic information theory, specifically algorithmic causality.
To perform this analysis, we assume the following, and state our justifications for such assumptions:
\begin{enumerate}
    \item The learning algorithm is implemented in on a universal Turing machine with a universal programming language.
    \paragraph{Justification:}
    \emph{This is a standard assumption in machine learning research that the machine learning algorithm can be implemented on a machine.}
    \item Each initial parameter of the learnable functions $\allfunc$ is jointly algorithmically independent of the other initial parameters.
    \paragraph{Justification:}
    \emph{This is a standard assumption in machine learning research that the noise from the random number generator is given background knowledge~\citep[\S2.3]{janzing2010causal}, thus allowing us to ignore possible dependencies among the parameters induced by the random number generator in developing our algorithms.}
    \item The function nodes of~\modelacronym~-- $\texttt{APPLY}$, $\texttt{UPDATE}$, and the internal function nodes of $\camech$ -- are $O(1)$-length programs.
    \paragraph{Justification:}
    \emph{
    Assuming \texttt{APPLY} (which encompasses the transition and reward functions of the MDP, see~\S\ref{sec:from_monolithic_policies_to_decision_mechanisms}) is an $O(1)$-length program is reasonable because it is a standard assumption that they are fixed with respect to the activations and parameters of the learning algorithm.
    Assuming \texttt{UPDATE} is an $O(1)$-length program is a standard assumption in machine learning research that the source code that computes the update rule (e.g. a gradient descent step) is agnostic to the feedback signals (e.g. gradients) it takes as input.
    Assuming the internal function nodes of $\camech$ are $O(1)$-length programs is reasonable because these function nodes are operations in the programming language like addition, multiplication, etc that are agnostic to the activations and parameters of the learning algorithm.
    }
    \item $\mathbb{L}$ is faithful. That is, any conditional independence among the data nodes ($\texttt{f}$, $x$, $\delta$, and the internal variable nodes of $\camech$) in $\mathbb{L}$ is due to the causal structure of $\mathbb{L}$ rather than a non-generic setting of these data nodes.
    \paragraph{Justification:}
    \emph{Faithfulness has been justified for standard causal models~\citep{10.5555/2074158.2074205}.
    Deriving an algorithmic analog has been the subject of ongoing work~\citep{lemeire2013replacing,lemeire2016conditional}.
    For our work, a violation of faithfulness means that two nodes $x$, $y$ in the computational graph have $I(x:y) \overset{+}{=} 0$ but are not $d$-separated in the computational graph.
    This would happen if $x$ and $y$ were tuned in such a way that makes one compressible given the other.
    Given assumption (2) above, the source of a violation of faithfulness must be the data experienced by the learning algorithm.
    Indeed, the data could be such that after learning certain parameters within $\allfunc$ may be conditionally independent given the training history, as suggested by~\citet{csordas2020neural,filan2021clusterability,watanabe2019interpreting}.
    However, as our focus is on theoretical results that hold regardless of the data distribution the learning algorithm is trained on, we consider the specific instances where the data does induce such faithfulness violations as ``non-generic'' and thus out of scope of the paper.
    }
\end{enumerate}

\section{Additional Theoretical Results} \label{appdx:additional_theoretical_results}
All modular credit assignment mechanisms must be factorized in the following way:
\begin{theorem}[\textbf{modular factorization}] \label{thm:modular_factorization}
The credit assignment mechanism $\camech(\bm{\tau}, \allfunc) \rightarrow \bm{\delta}$ is modular if and only if
\begin{align}
    K\left(\bm{\delta} \mgiven \trace, \allfunc\right) \overset{+}{=} \sum_{\traceindex=1}^\tracelen K\left(\delta_\traceindex \mgiven \trace, \allfunc\right).
\end{align}
\end{theorem}
For modular credit assignment mechanisms, the complexity of computing feedback for the entire system is minimal because all redundant information among the gradients has been ``squeezed out.''
This connection between simplicity and modularity is another way of understanding why if a credit assignment mechanisms were not modular it would be impossible for $\camech$ to modify a function without simultaneously modifying another, other than due to non-generic instances when $\delta_\traceindex$ has a simple description, i.e. $\delta_\traceindex = 0$, which, unless imposed, are not likely to hold  over all iterations of learning.

\section{Proofs} \label{appdx:proofs}
Given the assumptions stated in \S\ref{appdx:assumptions}, we now provide the proofs for our theoretical results.
We will prove Lemma~\ref{lem:algorithmic_causal_model_of_learning} first.
Together with the faithfulness postulate (Post.~\ref{post:faithfulness}) and the equivalence of algorithmic Markov conditions (Thm.~\ref{thm:equivalence_of_algorithmic_markov_conditions}) we can prove algorithmic independence by inspecting the graph of $\mathbb{L}$ for $d$-separation.

\subsection{Dynamic Modularity and the Algorithmic Causal Model of Learning}
\textbf{Lemma~\ref{lem:algorithmic_causal_model_of_learning} (algorithmic causal model of learning).}
\textit{Given a model of execution $\mexec$ and of credit assignment $\metamdp$, define the \textbf{algorithmic causal model of learning} (\modelacronym) as a dynamic computational graph $\acl$ of the learning process.
We assume $\camech$ has its own internal causal structure with internal variable and function nodes.
The function nodes of $\acl$ are \texttt{APPLY}, \texttt{UPDATE}, and internal function nodes of $\camech$.
The variable nodes of $\acl$ are $x$, \texttt{f}, $\delta$, and internal variable nodes of $\camech$.
\texttt{APPLY} and \texttt{UPDATE} are assumed to have length $O(1)$.
The internal function nodes of $\camech$ jointly independent, and along with the variable nodes of $\acl$, are assumed to not have length $O(1)$.
Then these variable nodes satisfy the algorithmic causal Markov condition with respect to $\acl$ for all steps of credit assignment.}
\begin{proof}
If we can express $\Pi$ in the form of the computational graph defined in Definition~\ref{def:computational_graph}, then we will have shown that $\acl$ is a well-defined computational graph, and so by Thm.~\ref{thm:algorithmic_causal_markov_condition} it satisfies the algorithmic causal Markov condition.
To express $\Pi$ as a computational graph, let us denote the internal function nodes $\Pi$ as $\texttt{g}^j$, indexed by $j$.
For a given input $u_j$ into $\texttt{g}^j$, we can equivalently write $\texttt{g}^j(u_j)$ as $\texttt{APPLY}(\texttt{g}^j, u_j)$.
Then since we assume $\texttt{APPLY}$ is $O(1)$, by treating $\texttt{g}^j$ as analogous to the auxiliary input $n_j$ in Definition~\ref{def:computational_graph}, $\Pi$ is a well-defined computational graph, which completes the proof.

\end{proof}
\begin{remark}
By Lemma~\ref{lemma:information_nonincrease}, if a set of data nodes in $\mathbb{L}$ are independent, then processing them separately with factor nodes of $\mathbb{L}$ will maintain this independence. For example, given that the \texttt{UPDATE} operation is applied in separately for each pair $(\texttt{f}^\funcindex, \sum_\traceindex \delta_\traceindex^\funcindex)$ to produce a corresponding $\texttt{f}^{\funcindex\prime}$, then if $(\texttt{f}^\funcindex, \sum_\traceindex \delta_\traceindex^\funcindex)$ were independent of $(\texttt{f}^\otherfuncindex, \sum_\traceindex \delta_\traceindex^\otherfuncindex)$ before applying \texttt{UPDATE}, then $\texttt{f}^{\funcindex\prime}$ would be independent of $\texttt{f}^{\otherfuncindex\prime}$ after applying \texttt{UPDATE}.
\end{remark}

\textbf{Theorem~\ref{thm:modular_credit_assignment} (modular credit assignment).}
\textit{Dynamic modularity is enforced at learning iteration $\learnindex$ if and only if static modularity holds at iteration $i=0$ and the credit assignment mechanism satisfies the modularity constraint.}
\begin{proof}
We will prove by induction on $\learnindex$.
The inductive step will make use of the equivalence between $d$-separation and conditional independence.

\textbf{Base case:} 
$\learnindex = 1$. There is no training history, so static modularity is equivalent to dynamic modularity.

\textbf{Inductive hypothesis:} 
Assuming that dynamic modularity holds if and only if static modularity holds at $\learnindex = 0$ and modular credit assignment holds for learning iteration $\learnindex-1$, dynamic modularity holds if and only if static modularity and modular credit assignment hold for learning iteration $\learnindex$.

\begin{figure}[t]
    \centering
    \includegraphics[width=0.3\textwidth]{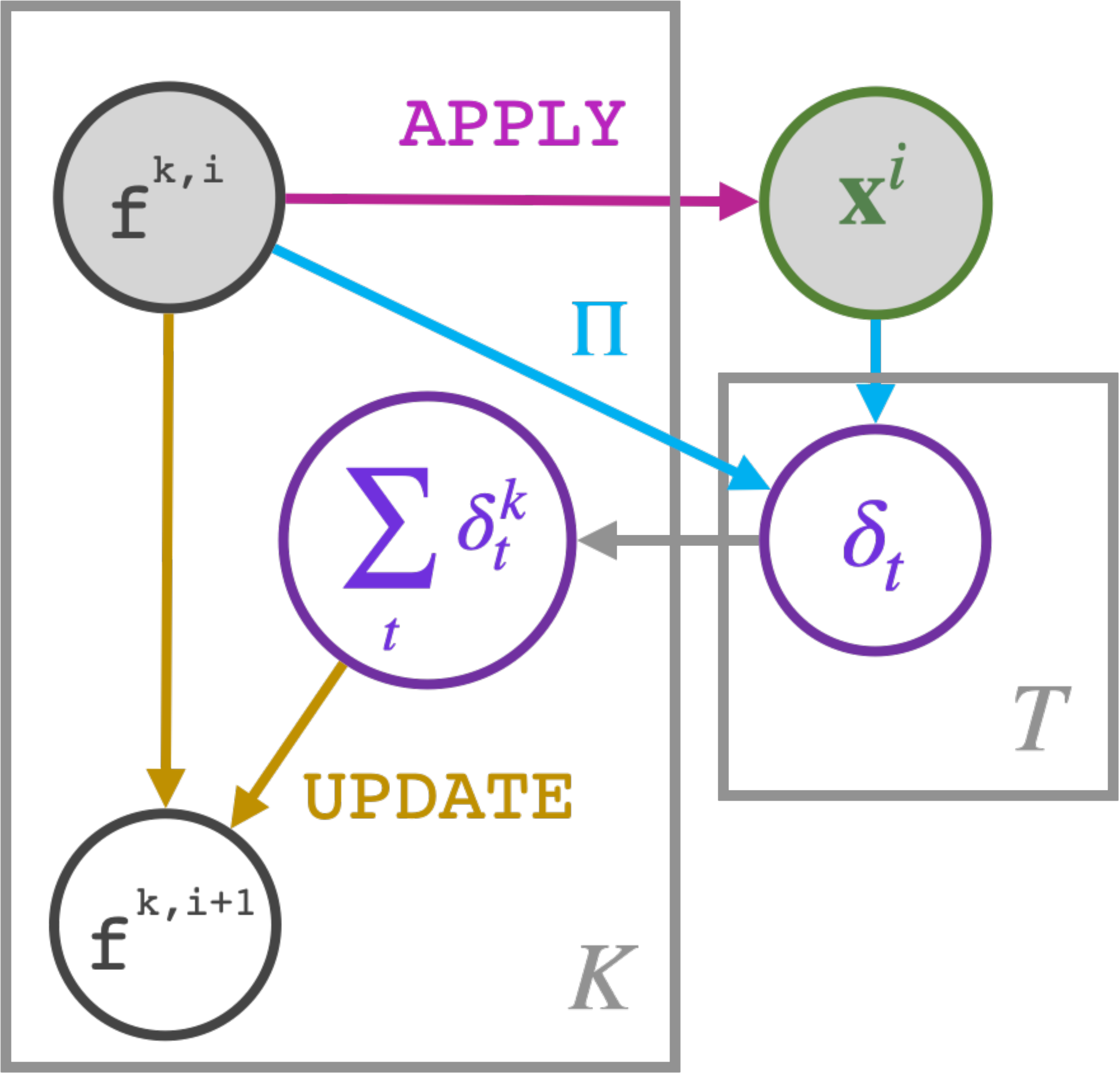}
    \caption{\small{This figure shows the computation graph of $\mathbb{L}$ across one credit assignment update.
    Inputs to the credit assignment mechanism are shaded.
    A modular credit assignment mechanism (shown with blue edges) is equivalent to showing the gradients $\delta_\traceindex$ as conditionally independent, as shown by the plate notation labeled with $\tracelen$.
    Dynamic modularity at iteration $\learnindex-1$ is equivalent to showing that the functions $\texttt{f}^{\funcindex, \learnindex}$ are inside the plate labeled with $\numfunc$.
    Then because the \texttt{UPDATE} operation, shown with yellow edges, operates only within the plate labeled with $\numfunc$, the updated functions $\texttt{f}^{\funcindex, \learnindex+1}$ are also conditionally independent given $(\trace, \allfunc)$.
    }}
    \label{fig:dynamic_modularity}
\end{figure}

\textbf{Inductive step:}
The modularity constraint states
\begin{align*}
    I\left(\delta_1, ..., \delta_M \mgiven \trace_\learnindex, \allfunc_\learnindex \right) &\overset{+}{=} 0.
\end{align*}
Dynamic modularity at iteration $\learnindex-1$ states that
\begin{align*}
    \forall \funcindex \neq \otherfuncindex, \quad I\left(\texttt{f}^{\funcindex, \learnindex} : \texttt{f}^{\otherfuncindex, \learnindex} \mgiven \trace_{\learnindex-1}, \allfunc_{\learnindex-1} \right) \overset{+}{=} 0.
\end{align*}
These two above statements correspond to the computational graph in Fig.~\ref{fig:dynamic_modularity}.
Note that by Def.~\ref{lemma:algorithmic_joint_conditional_independence}, disjoint subsets of $\delta_1, ..., \delta_\tracelen$ also have have zero mutual information up to an additive constant.
Letting these subsets be $\sum_\traceindex \delta_\traceindex^\funcindex$ where $\funcindex$ is the index of function $\thisfunc$ in $\allfunc$, then 
\begin{align} \label{eqn:independent_aggregated_feedback}
    I\left(\sum_\traceindex \delta_\traceindex^1, ..., \sum_\traceindex \delta_\traceindex^\numfunc \mgiven \trace_\learnindex, \allfunc_\learnindex \right) &\overset{+}{=} 0.
\end{align}
Then, as we can see by direct inspection in Fig.~\ref{fig:dynamic_modularity}, $\texttt{f}^{\funcindex_{\learnindex}}$ and $\texttt{f}^{\otherfuncindex_{\learnindex}}$ are $d$-separated by $(\trace_\learnindex, \allfunc_\learnindex)$, which is equivalent to saying that dynamic modularity holds for iteration $\learnindex$.
\end{proof}

\textbf{Theorem~\ref{thm:modularity_criterion} (modularity criterion).}
\textit{If $\acl$ is faithful, the modularity constraint holds if and only if for all $i$, outputs $\delta_\traceindex$ and $\delta_\othertraceindex$ of $\camech$ are $d$-separated by its inputs $\alldata$ and $\allfunc$.}
\begin{proof}
The forward direction holds by the equivalence of algorithmic causal Markov conditions (Thm.~\ref{thm:equivalence_of_algorithmic_markov_conditions}), and the backward direction holds by the faithfulness assumption.
\end{proof}

\textbf{Theorem~\ref{thm:modular_factorization} (modular factorization).}
\textit{The credit assignment mechanism $\camech(\bm{\tau}, \allfunc) \rightarrow \bm{\delta}$ is modular if and only if
\begin{align}
    K\left(\bm{\delta} \mgiven \trace, \allfunc\right) \overset{+}{=} \sum_{\traceindex=1}^\tracelen K\left(\delta_\traceindex \mgiven \trace, \allfunc\right).
\end{align}}
\begin{proof}
The proof comes from the definition of algorithmic mutual information (Def.~\ref{def:joint_conditional_algorithmic_mutual_information}).
\begin{align}
    K\left(\bm{\delta} \mgiven \trace, \allfunc\right) &\overset{+}{=} \sum_{\traceindex=1}^\tracelen K\left(\delta_\traceindex \mgiven \trace, \allfunc\right) \\
    \sum_{\traceindex=1}^\tracelen K\left(\delta_\traceindex \mgiven \trace, \allfunc\right) - K\left(\bm{\delta} \mgiven \trace, \allfunc\right) &\overset{+}{=} 0\\
    I\left(\delta_1, ..., \delta_\tracelen \mgiven \trace, \allfunc \right) &\overset{+}{=} 0.
\end{align}
\end{proof}

\subsection{Modularity in Reinforcement Learning}
In the following, we sometimes use $b_\traceindex$ instead of $b_{s_\traceindex}$ to reduce clutter.

\textbf{Corollary~\ref{cor:policy_gradient} (policy gradient).}
\textit{All policy gradient methods do not satisfy the modularity criterion.
}
\begin{proof}
It suffices to identify a single shared hidden variable that renders $\delta_1, ..., \delta_\tracelen$ not $d$-separated.
Computing the policy gradient includes the log probability of the policy as one of its terms.
Computing this log probability for any action involves the same normalization constant $\sum_\funcindex b^\funcindex$.
This normalization constant is a hidden variable that renders $\delta_1, ..., \delta_\tracelen$ not $d$-separated, as shown in Fig.~\ref{fig:policy_gradient}.
\begin{figure}[h]
    \centering
    \includegraphics[width=0.2\textwidth]{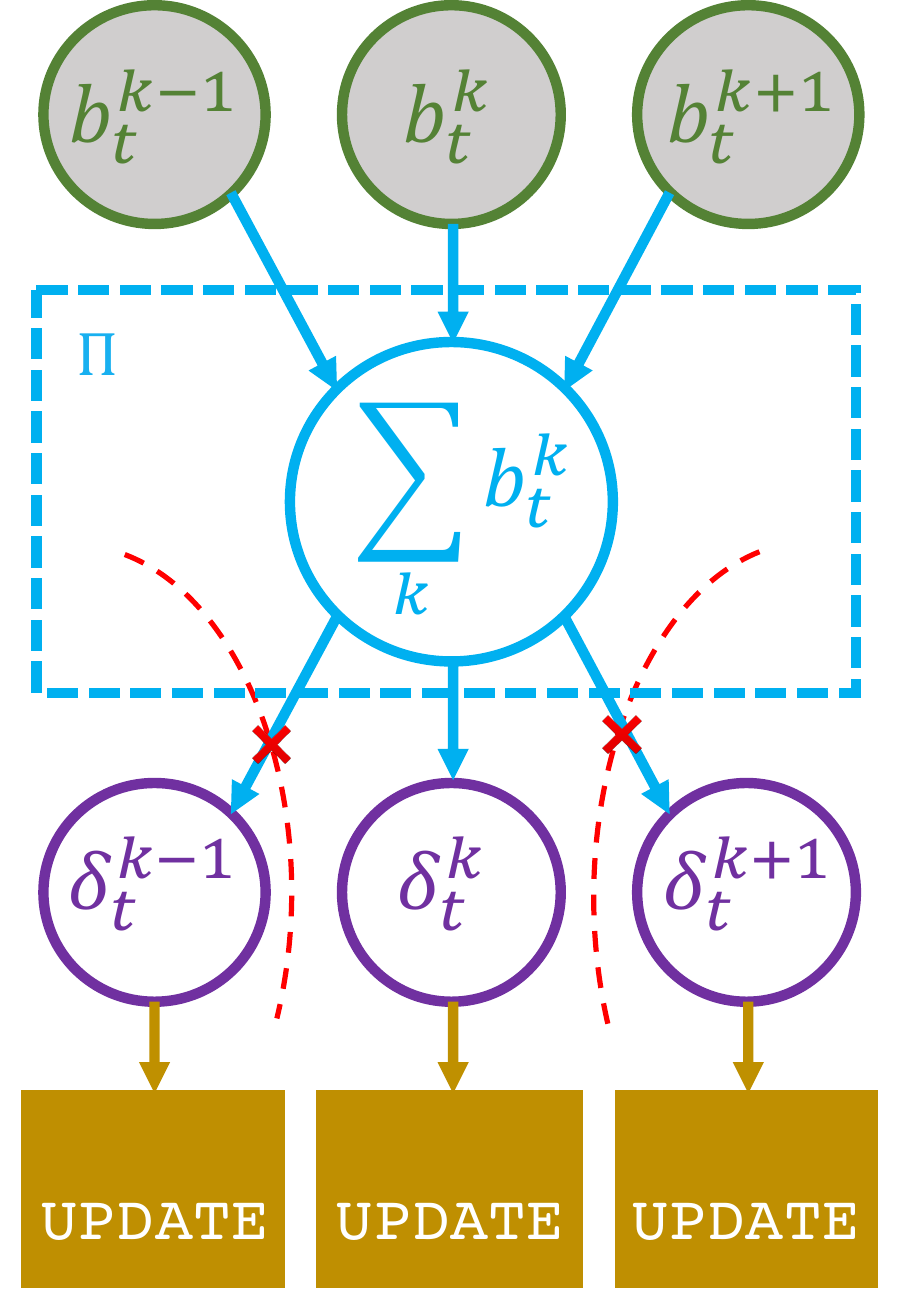}
    \caption{\small{
    This figure shows part of the computational graph within $\camech$ for policy gradient methods.
    Conditioning on $\trace$ implies we condition on the lightly shaded nodes. 
    $\sum_\funcindex b^\funcindex_t$ is the shared hidden variable that renders $\delta_1, ..., \delta_\tracelen$ not $d$-separated.}}
    \label{fig:policy_gradient}
\end{figure}
\end{proof}

\textbf{Corollary~\ref{cor:nsteptd} (n-step TD).}
\textit{All TD($n>1$) methods do not satisfy the modularity criterion.}
\begin{proof}
It suffices to identify a single shared hidden variable that renders $\delta_1, ..., \delta_\tracelen$ not $d$-separated.
TD($n > 1$) methods include a sum of estimated returns or advantages at different steps of the decision sequence that is shared among multiple $\delta_\traceindex$'s.
This sum is the hidden variable that renders $\delta_1, ..., \delta_\tracelen$ not $d$-separated, as shown in Fig.~\ref{fig:nsteptd}.
\begin{figure}[h]
    \centering
    \includegraphics[width=0.2\textwidth]{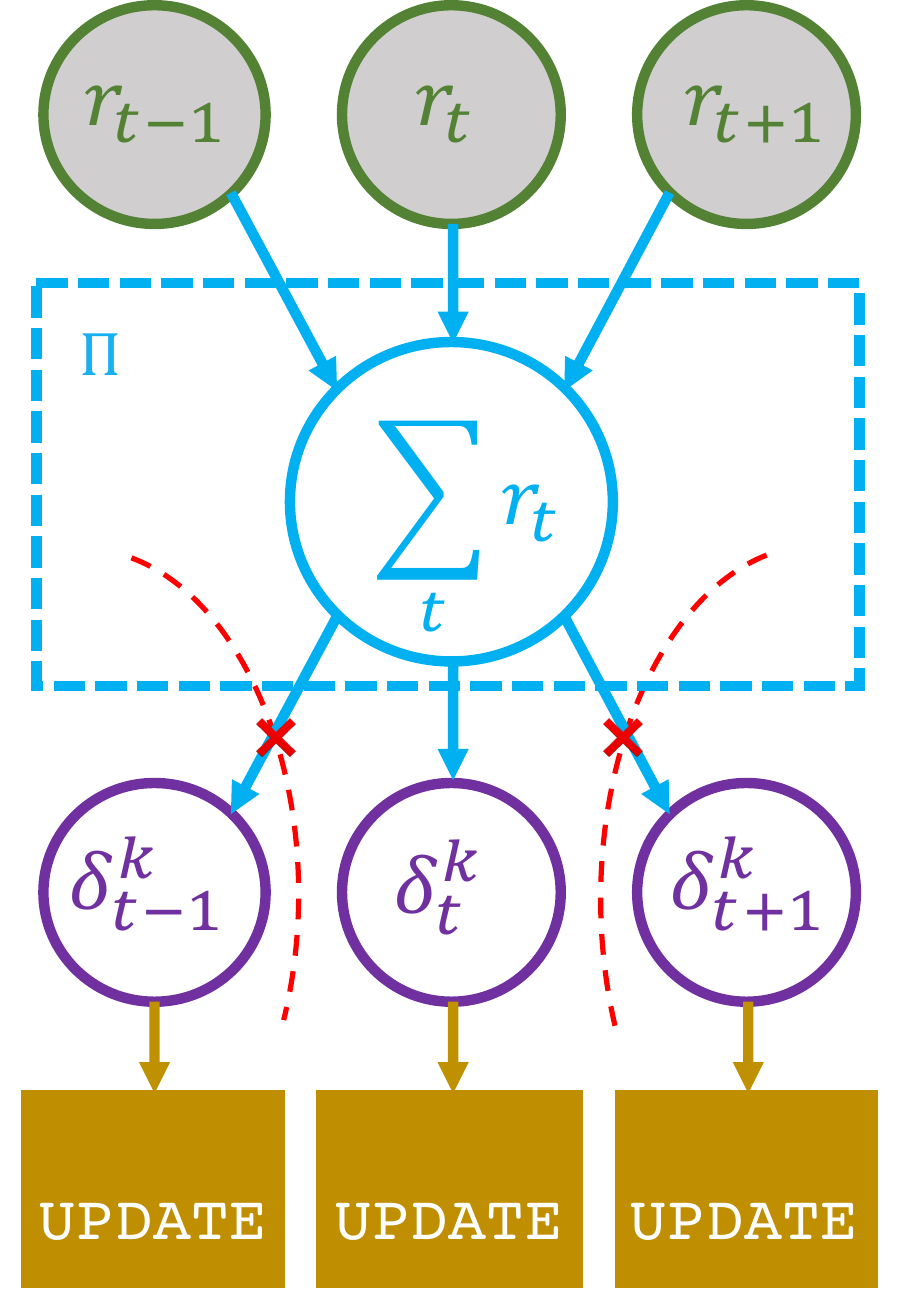}
    \caption{\small{
    This figure shows part of the computational graph within $\Pi$ for TD$(n>1)$ methods. 
    Conditioning on $\trace$ implies we condition on the lightly shaded nodes.  
    $\sum_t r_t$ is the shared hidden variable that renders $\delta_1, ..., \delta_\tracelen$ not $d$-separated.}}
    \label{fig:nsteptd}
\end{figure}
\end{proof}

\textbf{Corollary~\ref{cor:td0} (single-step TD).}
\textit{TD($0$) methods satisfy the modularity criterion for acyclic $\trace$.}
\begin{proof}
If the decision mechanism $\thisfunc$ were selected (i.e. $w_\traceindex^\funcindex = 1$) at step $i$, TD($0$) methods produce, for some function $g$, gradients as $\delta^\funcindex_\traceindex: = g(b^\funcindex_{\traceindex}, s_\traceindex, s_{\traceindex+1}, r_\traceindex, \allfunc)$.
Otherwise, $\delta^\funcindex_\traceindex := 0$.
The only hidden variable is $[\max_\otherfuncindex b^\otherfuncindex_{s_{\traceindex+1}}]$, and for acyclic $\trace$ there is only one state $s_\traceindex$ in $\trace$ that transitions into $s_{\traceindex+1}$.
Therefore the hidden variable is unique to each of $\delta_1, ..., \delta_\traceindex$, so $\delta_1, ..., \delta_\traceindex$ remain $d$-separated, as shown in Fig.~\ref{fig:td0}.
\begin{figure}[h]
    \centering
    \includegraphics[width=0.6\textwidth]{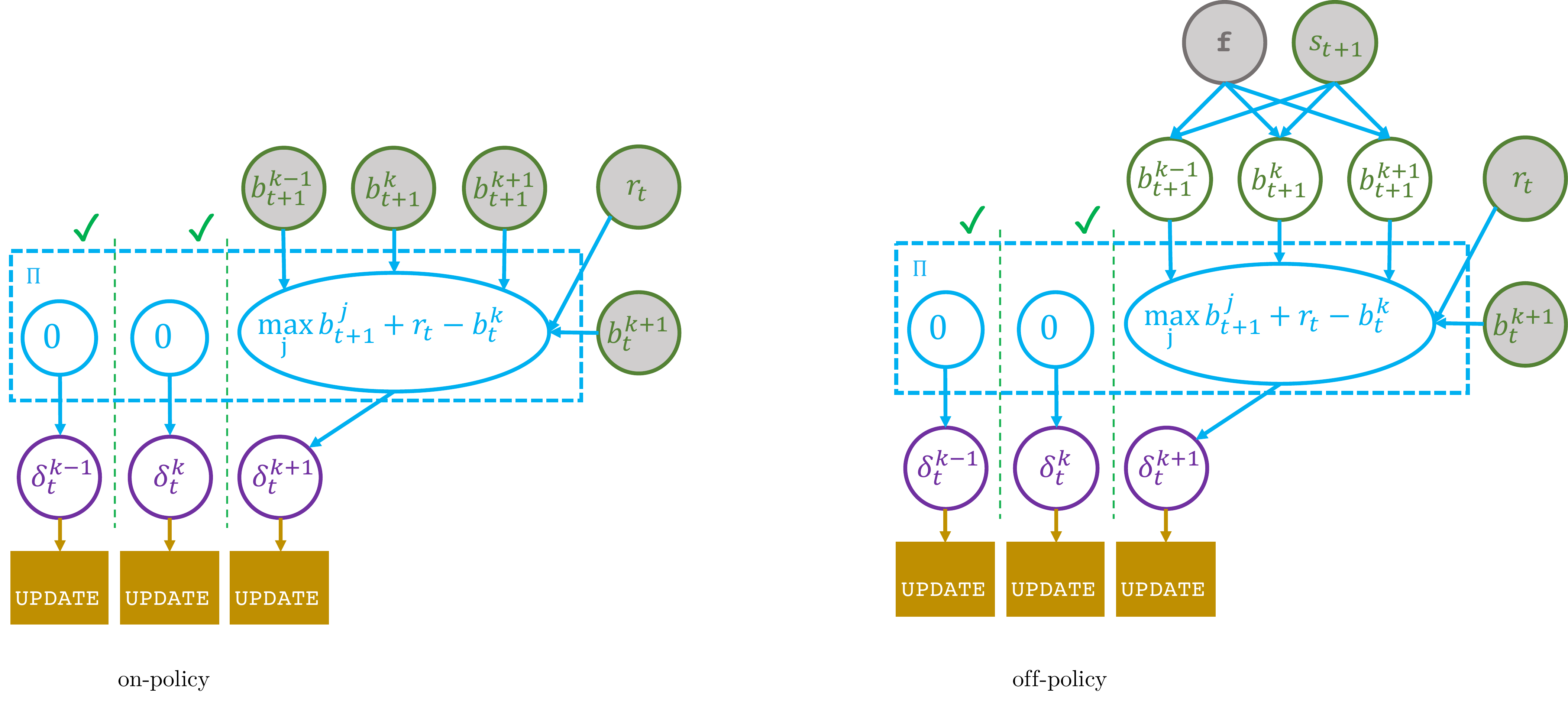}
    \caption{\small{
    This figure shows part of the computational graph within $\camech$ for on-policy and off-policy TD$(0)$ methods. 
    Conditioning on $(\trace, \allfunc)$ implies we condition on the lightly shaded nodes. 
    For on-policy methods such as~\clonedvickreysociety~and SARSA, the hidden variable would be $\max_\funcindex b_{t+1}^\funcindex$ for~\clonedvickreysociety~and the bid corresponding to the decision mechanism that was sampled through $\varepsilon$-greedy for SARSA.
    The figure shows $\max_\funcindex b_{t+1}^\funcindex$ for concreteness.
    For off-policy methods such as $Q$-learning, the bids $b_{t+1}$ are computed from $s_{t+1}$ and $\allfunc$, both of which we condition on.
    In both cases, the hidden variable is only parent to one of the $\delta_\traceindex$'s, and thus the $\delta_1, ..., \delta_\tracelen$ remain $d$-separated.
    }}
    \label{fig:td0}
\end{figure}
\end{proof}

\textbf{Corollary~\ref{cor:tabular} (tabular).}
\textit{In the tabular setting, Thm.~\ref{thm:modular_credit_assignment} holds for $Q$-learning, SARSA, and \clonedvickreysociety.}
\begin{proof}
In the tabular setting, decision mechanisms are columns of the $Q$-table corresponding to each action.
These columns do not share parameters, so static modularity holds.
Then because $Q$-learning, SARSA, and~\clonedvickreysociety~are TD($0$) methods, by Corollary~\ref{cor:td0}, their credit assignment mechanisms are modular.
Therefore Thm.~\ref{thm:modular_credit_assignment} holds.
\end{proof}

\textbf{Corollary~\ref{cor:funcapprox} (function approximation).}
\textit{In the function approximation setting, Thm.~\ref{thm:modular_credit_assignment} holds for TD($0$) methods whose decision mechanisms do not share parameters.}
\begin{proof}
The decision mechanisms of~\clonedvickreysociety~do not share weights, so static modularity holds.
By Corollary~\ref{cor:td0} its credit assignment mechanism is modular.
Therefore Thm.~\ref{thm:modular_credit_assignment} holds.
\end{proof}

\section{Simulation Details} \label{appdx:simulation_details}
We implemented our simulations using the PyTorch library~\citep{paszke2019pytorch}.

\subsection{Implementation Details} \label{appdx:implmentation_details}
\newcommand{\todo}{\textcolor{red}{TODO}}
The underlying PPO~\citep{schulman2017proximal} implementation used for~\clonedvickreysociety, PPO, and PPOF used a
policy learning rate of $4\times 10^{-5}$, 
a value function learning rate of $5\times 10^{-3}$, 
a clipping ratio of $0.2$, 
a GAE~\citep{schulman2015high} parameter of $0.95$,
a discount factor of $0.99$,
entropy coefficient of $0.1$,
and the Adam~\citep{kingma2014adam} optimizer.
For all algorithms, the policy and value functions for our algorithms were implemented as fully connected neural networks that used two hidden layers of dimension 20, with a ReLU activation.
All algorithms performed a PPO update every 4096 samples with a minibatch size of 256.

\subsection{Training Details} \label{appdx:training_details}
All learning curves are plotted from ten random seeds, with a different learning algorithm represented by a different hue.
The dark line represents the mean over the seeds.
The error bars represent one standard deviation above and below the mean.

Our protocol for transfer is as follows.
A transfer problem is defined by a (training, transfer) task pair, where the initial network parameters for the transfer task are the network parameters learned the training task for $H$ samples.
In our simulations, we set $H$ to $10^7$ because that was about double the number of samples for all algorithms to visually converge on the training task for all seeds. 
To calculate the relative sample efficiency of~\clonedvickreysociety~over PPO and PPOF (e.g. 13.9x and 6.1x respectively in the bottom-up right corner of Fig.~\ref{fig:task_graphs}), we set the criterion of convergence as the number of samples after which the return deviates by no more than $\varepsilon = 0.01$ from the optimal return for 30 epochs of training, where each epoch of training trains on 4096 samples.

\subsection{Environment Details} \label{appdx:environment_details}
The environment for our experiments shown in Figs.~\ref{fig:intervention_on_the_transition_function} and~\ref{fig:task_graphs} are represented as discrete-state, discrete-action MDPs.
Each state is represented by a binary-valued vector.

The structure of the MDP can best be explained via an analogy to a room navigation task, which we will explain in the context of the $A \rightarrow B \rightarrow C$ task in the \emph{linear chain} topology.
In this task, there are four rooms, room 0, room 1, room 2, and room 3.
Room 0 has two doors, labeled $A$ and $F$, that lead to room 1.
Room 1 has two doors, labeled $B$ and $E$, that lead to room 2.
Room 2 has two doors, labeled $C$ and $D$.
Doors are unlocked by keys.
The state representation is a concatenation of two one-hot vectors.
The first one-hot vector is of length four; the ``1'' indicates the room id.
The second one-hot vector is of length six; the ``1'' indicates the presence of a key for door $A$, $B$, $C$, $D$, $E$, or $F$.
Only one key is present in a room at any given time.
If the agent goes through the door corresponding to the key present in the room, then the agent transitions into the next room; otherwise the agent stays in the same room.
In the last room, if the agent opens the door corresponding to the key that is present in the room, then the agent receives a reward of $1$.
All other actions in every other state receive a reward of $0$.
Therefore the agent only gets a positive reward if it opens the correct sequence of doors.
For all of our experiments, the optimal policy is acyclic, but a suboptimal decision sequence could contain cycles.

Therefore, for the training task in the \emph{linear chain} topology where the optimal solution is $A \rightarrow B \rightarrow C$, the optimal sequence of states are
\begin{verbatim}
    [1, 0, 0, 0 ; 1, 0, 0, 0, 0, 0]  # room 0 with key for A
    [0, 1, 0, 0 ; 0, 1, 0, 0, 0, 0]  # room 1 with key for B
    [0, 0, 1, 0 ; 0, 0, 1, 0, 0, 0]  # room 2 with key for C.
\end{verbatim}
For the transfer task whose optimal solution is $A \rightarrow B \rightarrow D$, the optimal sequence of states are
\begin{verbatim}
    [1, 0, 0, 0 ; 1, 0, 0, 0, 0, 0]  # room 0 with key for A
    [0, 1, 0, 0 ; 0, 1, 0, 0, 0, 0]  # room 1 with key for B
    [0, 0, 1, 0 ; 0, 0, 0, 1, 0, 0]  # room 2 with key for D.
\end{verbatim}
Whereas for the \emph{linear chain} topology the length of the optimal solution is three actions, for the \emph{common ancestor} and \emph{common descendant} topologies this length is two actions.
\emph{Common ancestor} and \emph{common descendant} are multi-task problems.
As a concrete example, the training task for \emph{common ancestor} is a mixture of two tasks, one whose optimal solution is $A \rightarrow B$ and one whose optimal solution is $A \rightarrow C$.
Following the analogy to room navigation, this task is set up such that after having gone through door $A$, half the time there is a key for door $B$ and half there is a key for door $C$.

\subsection{Computing Details}
For our experiments, we used the following machines:
\begin{itemize}
    \item AWS: c5d.18xlarge instance
    \item Azure: Standard D64as\_v4 (64 vcpus, 256 GiB memory), 50GiB Standard SD attached.
\end{itemize} 
The average runtime training on $10^7$ samples was three hours for PPO and PPOF and 6 hours for~\clonedvickreysociety. 
\end{appendices}

\end{document}